\documentclass[runningheads]{llncs}
\usepackage{graphicx}

\usepackage{tikz}
\usepackage{comment}
\usepackage{amsmath,amssymb} 
\usepackage{color}
\usepackage[outline]{contour}
\usepackage{pifont}
\usepackage{multirow}

\usepackage{enumitem}
\setitemize{noitemsep,topsep=0pt,parsep=0pt,partopsep=0pt}

\newcommand{\modelName}{Dessurt}

\begin{document}
\pagestyle{headings}
\mainmatter
\def\ECCVSubNumber{6561}  

\title{End-to-end Document Recognition and Understanding with \modelName{}} 

\titlerunning{End-to-end Document Recognition and Understanding with \modelName{}}
%
\author{Brian Davis\inst{1} \and
Bryan Morse\inst{1} \and
Bryan Price\inst{2} \and 
Chris Tensmeyer\inst{2} \and
Curtis Wigington\inst{2} \and
Vlad Morariu\inst{2}}
\authorrunning{B. Davis et al.}
%
\institute{Brigham Young University, Provo UT, USA 
\email{\{briandavis,morse\}@byu.edu}
\and
Adobe Research, USA
\email{\{bprice,tensmeye,wigingto,morariu\}@adobe.com}}
\maketitle

\begin{abstract}
We introduce \modelName{}, a relatively simple document understanding transformer capable of being fine-tuned on a greater variety of document tasks than prior methods.
It receives a document image and task string  as input and generates arbitrary text autoregressively as output.
Because \modelName{} is an end-to-end architecture that performs text recognition in addition to the document understanding, it does not require an external recognition model as prior methods do. Dessurt is a more flexible model than prior methods and is able to handle a variety of document domains and tasks.
We show that this model is effective at 9 different dataset-task combinations. 

\end{abstract}

\section{Introduction}

Document understanding is an area of research attempting to automatically extract information from documents, whether that be specific key information, answers to natural language questions, or other similar elements. While there have been many approaches, the research community has begun to gravitate around pre-trained transformers as general purpose solutions. Beginning with LayoutLM~\cite{layoutlm}, these models began as BERT-like transformers incorporating  spatial/layout information and later visual features.
In general, we refer to these as the LayoutLM family. The LayoutLM family of models are pre-trained on a large corpus of document images and then fine-tuned to their particular tasks.

The LayoutLM family consists of encoder-only transformers, meaning predictions are only made for the input tokens. These state-of-the-art models are two-stage models, where text recognition is first performed by an external OCR model to obtain the input text tokens for the transformer. 
We see two limitations coming from these architecture choices:
\begin{enumerate}
    \item A limited output space, having predictions only for individual input tokens. While they can classify the input tokens, they cannot produce additional outputs, e.g., arbitrary text or token relationships, without additional submodules.
    \item Dependence on high quality external OCR text segmentation and recognition. Encoder-only transformers are incapable of inserting new tokens if the OCR missed or under-segmented text. A single incorrectly recognized character in an OCR'd word can cause a wrong word embedding to be used or cause the word to be out of vocabulary. Relatedly, discrete input tokens lack the uncertainty the text recognition model may have in its predictions. For clean, modern documents, this generally isn't an issue as the OCR models used are quite robust. However, for handwritten or degraded historical documents, OCR quality can be poor and lead to prediction errors.
\end{enumerate}

\begin{figure}[t!]
\centering
\includegraphics[width=0.99\textwidth]{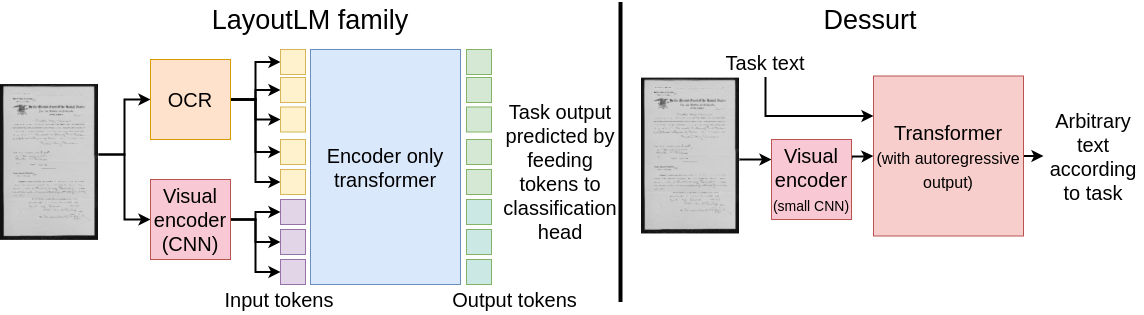}
\caption{The LayoutLM family of document transformers require OCR and output is tied to the tokens. \modelName{} does not require any separate models and can generate arbitrary text to solve a variety of tasks.}
\label{fig:compare}
\end{figure}

\contourlength{0.15pt}
\contournumber{8}%
To combat these flaws we introduce \textbf{\modelName{}}: \contour{black}{D}ocument \contour{black}{e}nd-to-end \contour{black}{s}elf-\contour{black}{s}upervised \contour{black}{u}nderstanding and \contour{black}{r}ecognition \contour{black}{t}ransformer. 
\modelName{} is a novel, general document understanding architecture that can perform a great variety of document tasks. \modelName{} operates in an end-to-end manner with a single pass:  text segmentation and recognition are learned implicitly. \modelName{} takes only the image and task text as input and can auto-regressively produce arbitrary text as output. Fig.~\ref{fig:compare} compares \modelName{} to the LayoutLM family at a high level architecturally. The first limitation of the LayoutLM family is easily solved with \modelName{}'s auto-regressive output. Because text recognition is implicit, rather than provided as explicit OCR results, \modelName{} is able to resolve text recognition uncertainty or ambiguity in a task-focused way. Additionally, the auto-regresssive output decouples \modelName{}'s output from the text recognition. These together address the second limitation.  See Table~\ref{tbl:compare} for a comparison of architecture features. 

\begin{table}[t]
\caption{Model class capabilities }
\label{tbl:compare}
\begin{tabular}{l|c|c|c}
                & Handwriting  & Arbitrary output & Apply to different visual domain \\ \hline
LayoutLM family & OCR dependant &   \ding{55}            & Fine-tune two models                           \\
\modelName{}   & \checkmark    & \checkmark       & Fine-tune single model                        
\end{tabular}
\end{table}

Because \modelName{} takes both an input image and text and can output any arbitrary text, it can complete a greater variety of tasks compared to the LayoutLM family of transformers. Particularly we solve a form parsing task (form image to JSON) that the LayoutLM family cannot handle without additional modules.  
Also, when retraining for a different visual domain, \modelName{}'s simple end-to-end design means only one model needs to be fine-tuned.  For the LayoutLM family, both the recognition and transformer models would need to be fine-tuned.

Like prior methods, we pre-train on the IIT-CDIP dataset~\cite{iit}, a large collection of document images, with a masked language modeling task. We also introduce three synthetic document datasets to better capture natural language, structured documents, and handwriting recognition. Finally, we introduce new pre-training tasks to teach \modelName{} to read and located text, and to parse structured documents.

We validate our claims of \modelName{}'s flexibility by applying it to six different document datasets across six different tasks:
1) Document question answering, with both DocVQA~\cite{docvqa} and HW-SQuAD~\cite{hwsquad}, 2) Form understanding and 3) Form parsing, with both the FUNSD~\cite{funsd} and NAF~\cite{davis} datasets, 4) Full-page handwriting recognition and 5) Named entity recognition on the IAM handwriting database, and 6) Document classification with the RVL-CDIP dataset.  
Of particular interest, both NAF and IAM datasets require handwriting recognition, the NAF being comprised of difficult historical documents. These are domains in which the LayoutLM family would need 
to fine-tune its recognition model as well,
but \modelName{} can fine-tune on without adjustments. We note that \modelName{} does not achieve state-of-the-art results on the most tasks evaluated, but it is capable of operating on a larger range of tasks than individual state-of-the-art models.

In summary, our primary contributions are
\begin{itemize}
    \item \modelName{}, a novel, general document understanding architecture capable of both performing text recognition and document understanding in an end-to-end manner and producing arbitrary text output,
    \item A collection of synthetic datasets and tasks for pre-training an end-to-end document understanding model for a variety of possible final tasks,
    \item An evaluation of \modelName{} fine-tuned on 9 dataset-task combinations, and
    \item Our code, pre-trained model, and datasets which will be made available at \url{https://github.com/herobd/dessurt}
\end{itemize}





\section{Related Work}

\subsection{LayoutLM Family}
Document understanding has become largely dominated by transformer architectures. Beginning with LayoutLM(v1)~\cite{layoutlm} the goal was to bring the success of transformers like BERT~\cite{bert} in the natural language space into the more visual domain of documents. LayoutLM pre-trained in a very similar manner to BERT, but included 2D spatial position information. 

BROS~\cite{bros2021}, TILT~\cite{tilt},  and LayoutLMv2~\cite{layoutlmv2} improved the architecture by introducing spatially biased attention, making the spatial information even more influencial. LayoutLMv2 also introduced visual tokens as many layout cues are captured more visually than spatially.

Visual tokens can be overshadowed by textual tokens. In an effort to make the visual processing more important, DocFormer~\cite{docformer} forced feature updates to be from both textual and visual features.

We note that TILT and DocFormer use only visual features extracted near the text tokens spatially, making them blind to areas of the form without text. LayoutLMv2 extracts visual tokens across the entire document.

\subsection{End-to-end Models}

Models in the LayoutLM family have been evaluated without taking text recognition into account. Many document understanding datasets come with pre-computed OCR results used by everyone. While this is useful in making comparisons, text recognition is an essential task and for visually difficult documents can become a challange in itself.

One aim of an end-to-end method can be to accomplish both recognition and understanding in one pass. Another aim might be to learn the output text in a manner that allows arbitrary output predictions.
DocReader~\cite{docreader} is an end-to-end method for key information extraction. While it does rely on external OCR, it uses an RNN to predict arbitrary text.

We note that a concurrent pre-print work on end-to-end document understanding, Donut~\cite{donut}, has been introduced, and shares an architecture similar in design to \modelName{}. It also utilizes a Swin~\cite{swin} encoder but uses a BART-like~\cite{bart} decoder. Donut differs from \modelName{} primarily in how the cross attention occurs and in pre-training.  Donut shares many of the same advantages of \modelName{}.

\section{Model}

\begin{figure}[t]
\centering
\includegraphics[width=0.99\textwidth]{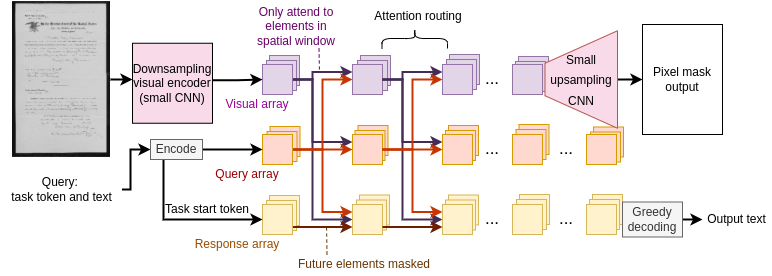}
\caption{\modelName{} architecture}
\label{fig:model}
\end{figure}

\modelName{} is a novel end-to-end framework for handling general document understanding problems. It takes as input an image and query text, and outputs arbitrary text appropriate for the given tasks. It handles character recognition, including handwriting recognition, implicitly as part of the network.

The architecture is shown in Fig~\ref{fig:model}. The model processes three streams of tokens: 1) Visual tokens that encode visual information about the document image, 2) Query tokens that encode the task the model is to perform, and 3) Autoregressive response tokens where the output response is formed. The model progresses through three main stages: Input encoding, cross-attention, and output decoding.

\emph{Input encoding: \ }   The input consists of an image and a query token string. Because the Swin~\cite{swin} layers we use require a fixed size image input, we use an input image size of 1152 $\times$ 768. This large size is needed as we process an entire page at once and must ensure small text is legible.
The input image is 2-channeled, one being the grayscale document, the other being a highlight mask used in some tasks.
The query tokens begin with a special task token indicating the desired task and then potentially have some text providing context for the task (e.g., the question text).  
The response tokens are initialized with a task specific start token and during training contain the previous ground truth token for teacher-forcing.

The first step of the model is to encode the inputs into feature arrays to initialize the three streams. The input image is tokenized by passing it through a small downsampling CNN and adding learned 2D spatial embeddings. The input query text and response text are tokenized using the same process as BART ~\cite{bart} with standard sinusoidal position encoding.  These feature arrays are then passed to their respective token streams.

Note that the model does not require as input any OCR tokens corresponding to the image. The network implicitly recognizes the text. 

\emph{Cross-attention: \ }  The three streams  then pass through a series of cross-attention layers to allow them to share information and transfer that information into the response. The visual array is processed by Swin~\cite{swin} layers modified to not only attend to the other elements in the local window but also the query array. (We note that the biased attention remains for the visual elements.) The query array has standard Transformer~\cite{transformer} attention, but attends to the entire visual array in addition to the query array. 
The response array has standard autoregressive attention to previous response elements but also attends to the visual and query arrays. The arrays pass through series of eight of their respective cross-attention layers. The last two layers of the model update only the query and response arrays, with both layers attending to the final visual features.

\emph{Output decoding: \ } The final response array is decoded into text using greedy search decoding (where the most likely token is selected at each step), allowing it to predict text not found in the document. Additionally, we also output a pixel mask for use in training. This is produced by a small upsampling network using six transpose convolutions that process the final visual features.

 Specific implementation details for the model and its layers can be found in the accompanying Supplementary Materials.

\section{Pre-training Procedure}

\begin{figure}[t!]
\centering
\includegraphics[width=0.99\textwidth,trim=0 100 0 0, clip]{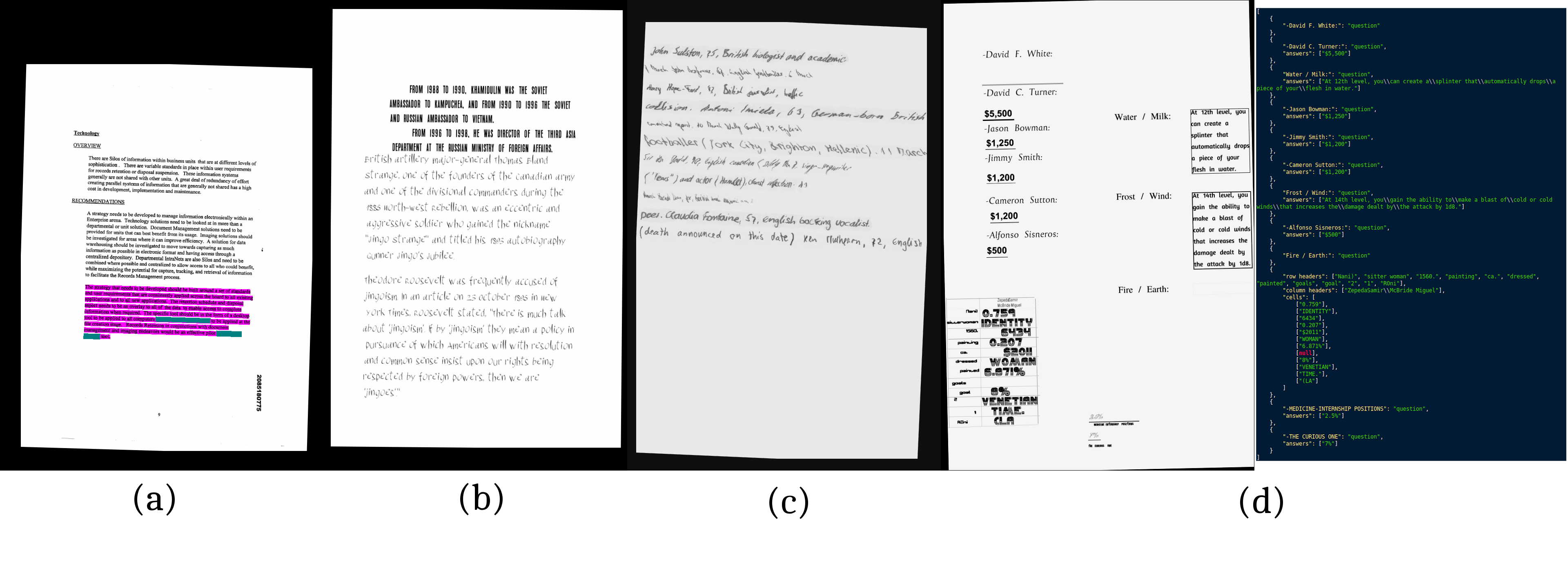}
\caption{Examples of data used in pre-training. (a) IIT-CDIP dataset image with Text Infilling task highlighting channel: highlight is magenta (value of 1), removed text is turquoise (value of -1). (b) Synthetic Wikipedia text. (c) Synthetic handwriting. (d) Synthetic form and its parse JSON.}
\label{fig:pretrain}
\end{figure}

The goal of the pre-training is to teach \modelName{}  to perform text recognition and document understanding and to have general language model capabilities like BERT. We pretrain several datasets with each dataset having multiple tasks associated with it. An example from each dataset is in Fig.~\ref{fig:pretrain}.

\subsection{IIT-CDIP dataset}
The IIT-CDIP dataset~\cite{iit} is a pre-training dataset used by several other document understanding transformers~\cite{layoutlm,layoutlmv2,docformer}. The OCR method we applied to the IIT-CDIP dataset is in the Supplementary materials.


There are several tasks defined with this dataset all of which are described in the Supplementary Materials. For brevity, we only describe the most important ones here.
The primary task (occurring 66\% of the time) is a Text Infilling task. It is a masked language modeling task inspired by the text infilling used to train BART~\cite{bart}; instead of replacing the removed text with a blank token, we delete them from the image. 
The entire block of text and the deleted areas are marked (with different values) in the input highlight channel, as seen in Fig.~\ref{fig:pretrain} (a).
The model then must predict the text of the entire block, filling in the deleted regions.  We also do a variant of this task where a single word is blanked from the image and the model must predict that single word.
There are several reading based tasks as well, such as to read on from the text provided in the query.

\subsection{Synthetic Wikipedia}
We want our pre-training to help the model understand natural language; however, the IIT-CDIP dataset only represents a skewed slice of natural language. 
Additionally, it represents a limited range of font styles. We choose to create an on-the-fly dataset by selecting random text from Wikipedia\footnote{\url{https://huggingface.co/datasets/wikipedia}}~\cite{wikidump} and rendering it as paragraphs in random locations with random fonts.

We pick a random article, random column width, random font, random text height, and random spacing (between word and new line). We render the words using the font and text height. We place the words in column/paragraph form, adjusting the column width to fit as much of the article as possible. 
We find blank space in the image where the paragraph can be added.
If one is found the paragraph is added and we attempt to add another paragraph; otherwise, the image is complete.
An example generated image is seen in Fig.~\ref{fig:pretrain} (b). 

To obtain our font database, we scrape all the free-for-commercial-use fonts from 1001fonts.com, giving us a set of over 10,000 fonts. 
The script we used to scrape the fonts will be made available. More details on these fonts and our synthetic dataset creation are found in the Supplementary Materials

This dataset uses the same distribution of tasks as the IIT-CDIP dataset.

\subsection{Synthetic Handwriting}
\modelName{} must be able handle handwriting as several document understanding tasks require this. The IIT-CDIP dataset contains little handwriting and while our font database has ``handwritten'' fonts, they do not capture the variation present in real handwriting. There is, unfortunately, not a publicly available dataset of handwriting comparable in size to the IIT-CDIP datset. The IAM handwriting database~\cite{iam} is frequently used, but with fewer than 800 instances to train on, an autoregressive transformer could overfit during pre-training.

We choose instead to use synthetic handwriting. This allows us to generate a larger breadth of text, but at the cost of realism.
We use the full line handwriting synthesis method of Davis et al.~\cite{line_gen} to 
generate 800,000 lines of sequential text from Wikipedia articles, with a randomly sampled style 
for each line. We compose a document by sampling a random number of consecutive handwriting lines (to maintain language flow), selecting a random text height, random newline height, and random starting location on the page, and then placing the lines in the document in block/paragraph style.
We additionally apply warp grid augmentation~\cite{curtis} to each line to further add to the visual variation. An example image can be seen in Fig.~\ref{fig:pretrain} (c).

For the learning task, the model must read the entire page of handwriting. 

\subsection{Synthetic Forms}
We want \modelName{} to be capable of extracting the information from highly structured documents, but given the lack of structured information present in our IIT-CDIP annotations, we decided to generate synthetic forms with known structure. The structure is based on the annotations of the FUNSD~\cite{funsd} dataset, which is primarily label-value pairs (or question-answer pairs) which are occasionally grouped under headers. We also include tables. 

To come up with label-value pairs, we use GPT-2~\cite{gpt2} to generate synthetic ``forms''. We give GPT-2 a prompt text (e.g.,~``This form has been filled out.'') followed by an example label-value pair, newline and a label with colon (e.g.,~``Date: 23 Mar 1999{\textbackslash}nName:''). GPT-2 then usually generate a series of label-value pairs separated by colons and newlines, which is easily parsed.
All the label-value pairs from one generation are a label-value set in our dataset.
We sometimes use Wikipedia article titles as part of the prompt (e.g.~``This form should be filled out regarding \underline{Marvel Universe}'') which then become the header for that label-value set. We reuse previously generated labels and label-value pairs as new form prompts. 
The quality of GPT-2 output is limited, but we hope it reflects at least some of the semantics of label-value relationships.

The data for tables is more random. The row and column headers are random 1 to 3 word snippets from Wikipedia. A cell value is either a random number (with various formatting) or a random word.

A document is composed by randomly placing label-value sets and tables until a placement fails due to there not being enough room.
Some cells and values are blanked.
More details on the form generation process can be found in the Supplementary Materials.

The primary task on the forms (occurring about half the time) is to parse it into a JSON capturing the text and structure of the form. An example synthetic form and its corresponding JSON are seen in Fig.~\ref{fig:pretrain} (d).
We also have tasks where the query has an entity on the form and the model must predict the class of the entity and then read the entities it is linked to. To ensure an understanding of tables, there are also table-specific tasks such as retrieving a cell based on a query with the row and column header, or listing all the row/column headers for a table.
All the tasks used are described in the Supplementary Materials.

\subsection{Distillation}
Because \modelName{} has a unique architecture, we could not use pre-trained transformer weights to initialize our model (like Donut~\cite{donut} or models in the LayoutLM family). This is clearly a disadvantage, so we attempt to infuse pre-trained language knowledge into \modelName{} in a different way: cross-domain distillation. We feed text to a pre-trained transformer teacher, and then render that text in a document image to pass to the student, \modelName{}. Then we apply the standard distillation loss introduced by Hinton et al.~\cite{distillation}, which guides the logit predictions of the student to match the teachers logits (the ``dark knowledge'').

Distillation is generally applied with a student and teacher getting the exact same inputs.
We are attempting something fairly unique which is to apply distillation across domains, textual to visual. 

To ensure architectural similarity, we need the teacher to be an autoregressive model. 
For this we use BART, an encoder-decoder transformer where the decoder is an autoregressive model with cross attention to the encoder (a vanilla transformer encoder). Both BART and \modelName{} will be given the Text Infilling task which BART was pre-trained with. BART gets the masked text as input to its encoder, and \modelName{} gets the rendered text with deleted regions as input (and the query token indicating the Test Infilling task) and then they both autoregressively output the input text with the blanks filled in. 

The token probabilities \modelName{} predicts for a blanked region reflect not only its language modeling, but also the uncertainty it has in reading the other words on the page. 
For BART the probabilities are only the language modeling; it has no uncertainty about reading. 
We minimize the reading uncertainty \modelName{} has when performing distillation by selecting a subset of ``easy'' fonts and reducing the variability with which the documents are rendered. 
More details on this are in the Supplementary Materials.

\subsection{Training}

We employ a simple curriculum to to prioritize certain aspects during early training. This is due to the need for recognition to be learned (to a certain degree) before the understanding tasks can be solved and the difficulty of learning recognition on dense multi-line documents in a semi-supervised fashion.

We first train \modelName{} on small images (96 $\times$ 384) of synthetic Wikipedia text with simple reading tasks 
for 150,000 iterations. Not only is the visual space small, but the output sequence length is short. 
We then use full-sized synthetic Wikipedia text documents for 200,000 iterations with primarily reading tasks. Finally, the model enters normal pre-training.

The iterations we outline here are what were used for the ablation models. For our primary evaluation we use a model that was pre-trained in total for over 10 million iterations (~110k weight update steps) during development (meaning datasets and tasks were added throughout the training), but followed roughly the same curriculum. The ablation models were pre-trained for 1 million iterations with all datasets and tasks being introduced at once.

We used data parallelism over 6 Nvidia Tesla P100s, which can each only hold a batch size of 1. We use gradient accumulation of 128 iterations, leading to an effective batch size of 768, with approximately  7,800 weight update steps for the ablation models. The last 4 million iterations of the final modal used gradient accumulation of 64 iterations, meaning the effective batch size was 384. We use the AdamW optimizer~\cite{adamw} and a learning rate of $10^{-4}$ and weight decay of 0.01.




\section{Experiments}

\begin{table}[t]
\caption{A summary of the end tasks we use to evaluate \modelName{} and their attributes. The term "special output" refers to whether the tasks requires more than standard token prediction employed by most in the LayoutLM family}\label{tab:datasets}
\begin{tabular}{l|l|l|l|l|l}
Dataset &
  Task &
  Domain &
  \begin{tabular}[c]{@{}l@{}}Requires\\ handwriting\\ recognition\end{tabular} &
  \begin{tabular}[c]{@{}l@{}}Requires\\ special\\ output\end{tabular} &
  \begin{tabular}[c]{@{}l@{}}Train\\ set  \\size\end{tabular} \\ \hline
RVL-CDIP~\cite{rvl} &
  Classification &
  Modern printed &
  No &
  No &
  320K\\ \hline
DocVQA~\cite{docvqa} &
  Question answering &
  Modern &
  Occasionally &
  No &
  39K\\ \hline
HW-SQuAD~\cite{hwsquad} &
  Question answering &
  \begin{tabular}[c]{@{}l@{}}Synthetic\\ handwriting\end{tabular} &
  Yes (easier) &
  No &
  68K \\ \hline
\begin{tabular}[c]{@{}l@{}}FUNSD~\cite{funsd}\end{tabular}&
  \begin{tabular}[c]{@{}l@{}}Entity / Relationship \\ detection\end{tabular} &
  \begin{tabular}[c]{@{}l@{}}Modern printed\\ forms\end{tabular} &
  No &
  No/Yes &
  130 \\ \hline
\begin{tabular}[c]{@{}l@{}}FUNSD\end{tabular} &
  Form parsing &
  \begin{tabular}[c]{@{}l@{}}Modern printed\\ forms\end{tabular} &
  No &
  Yes &
  130 \\ \hline
\begin{tabular}[c]{@{}l@{}}NAF~\cite{davis}\end{tabular} &
  \begin{tabular}[c]{@{}l@{}}Line / Relationship \\ detection\end{tabular} &
  Historic forms &
  Yes &
  No/Yes &
  921 \\ \hline
\begin{tabular}[c]{@{}l@{}}NAF\end{tabular} &
  Form parsing &
  Historic forms &
  Yes &
  Yes &
  921 \\ \hline
\begin{tabular}[c]{@{}l@{}}IAM~\cite{iam}\end{tabular} &
  Full page recognition &
  Handwriting &
  Yes &
  Yes &
  747\\ \hline
IAM NER~\cite{tuselmann} &
  \begin{tabular}[c]{@{}l@{}}Named entity\\ recognition\end{tabular} &
  Handwriting &
  Yes &
  No &
  747
\end{tabular}
\end{table}

To demonstrate the flexibility of \modelName{}, we evaluate it on the six document datasets and six diverse tasks
listed in Table~\ref{tab:datasets}. The RVL-CDIP dataset~\cite{rvl}  is a page classification dataset, which requires understanding overall layout and text topics. The DocVQA dataset~\cite{docvqa} requires both reading and layout comprehension. HW-SQuAD~\cite{hwsquad} is more focused on reading comprehension, but has difficult text (synthetic handwriting) to recognize. Both the FUNSD~\cite{funsd} and NAF~\cite{davis} datasets require form understanding, with a focus on label-value pairs in forms. The FUNSD dataset includes modern business documents, but the NAF dataset is uniquely challenging because it contains historical records with a both printed and handwritten text.
We take a task from Tüselmann et al.~\cite{tuselmann}, specifically named entity recognition over the IAM handwriting database (IAM NER), requiring both handwriting recognition and NLP capabilities. We also evaluate full-page handwriting recognition on the IAM database~\cite{iam}.
Each of these, and our experimental protocol for them, are discussed in more detail in the Supplementary Materials.
We also present an ablation study at the end of this section.

\subsection{RVL-CDIP}

We compare \modelName{} to several other models in Table~\ref{tab:rvl+docvqa} on document classification with the RVL-CDIP dataset. \modelName{} performs slightly below the state-of-the-art, but is comparable to the other models. We note that this problem requires a holistic view of the document and is likely benefiting from a strong vision model. We note that while \modelName{} uses a Swin architecture, it is shallower and narrower than the one used by Donut.

\begin{table}[]
\centering
\caption{Results on RVL-CDIP and DocVQA datasets}\label{tab:rvl+docvqa}
\begin{tabular}{l|c|l|l|l}
 & use OCR & \# params & \begin{tabular}[c]{@{}l@{}}RVL-CDIP\\ accuracy\end{tabular} & \begin{tabular}[c]{@{}l@{}}DocVQA\\ ANLS\end{tabular} \\ \hline
BERT\textsubscript{BASE}~\cite{bert}                  & \checkmark & 110M +OCR  & 89.8 & 63.5 \\
LayoutLM\textsubscript{BASE} (w/ img)~\cite{layoutlm} & \checkmark & 160M + OCR & 94.4 & -    \\
LayoutLM\textsubscript{BASE}~\cite{layoutlm}          & \checkmark & 113M + OCR & -    & 69.8 \\
LayoutLMv2\textsubscript{BASE}~\cite{layoutlmv2}        & \checkmark & 200M + OCR & 95.3 & 78.1\\
LayoutLMv2\textsubscript{BASE} w/ Tesseract OCR    & \checkmark & 200M + OCR & - & 48.2\\
DocFormer\textsubscript{BASE}~\cite{docformer}         & \checkmark & 183M + OCR & \textbf{96.2} & - \\
TILT\textsubscript{BASE}~\cite{tilt}                & \checkmark & 230M + OCR & 93.5    & \textbf{83.9} \\
Donut~\cite{donut}                                 &            & 156M       & 94.5 & 47.1 \\
Donut +10k trainset images~\cite{donut}              &            & 156M       & -    & 53.1 \\
\modelName{} (ours)                   &            & 127M       & 93.6 &     63.2
\end{tabular}
\end{table}


\subsection{DocVQA and HW-SQuAD}

For DocVQA, the model must locate the text that answers a textual question. The results are presented in Table~\ref{tab:rvl+docvqa}  with ANLS, a text edit-distance metric that accounts for multiple correct answers. Unlike RVL-CDIP, understanding the text in DocVQA is critical, likely leading to both \modelName{}'s and Donut's comparatively limited performance. Other models rely on strong external recognition methods; LayoutLMv2's performance significantly drops when using a weaker OCR.
\modelName{} outperforms Donut, likely due to its language-focused tasks and real data in pre-training. \modelName{}'s weakest areas for DocVQA are Figures/Diagrams and Image/Photo. This makes sense because the pre-training datasets are almost exclusively textual. 


 The HW-SQuAD dataset~\cite{hwsquad} is the popular question answering benchmark SQuAD~\cite{squad} rendered with handwritten fonts and noise. We evaluate on the task of machine comprehension, where the single document containing the answer is fed to the model. Unfortunately, the only prior method on this (\cite{recfree}) was doing text snippet retrieval, not question answering, and so is incomparable. We use ANLS as our metric as it seems well suited to the task and achieve 55.5\%.



\subsection{FUNSD and NAF}

Form parsing is the most difficult task we tackle, particularly with the NAF dataset, which is comprised of historical forms containing a good deal of handwriting.
In our full form parsing task the model must reproduce the entire contents of the form in a structured manner, including recognition of text. We have the model predict JSON using the same format used in pre-training (Fig.~\ref{fig:pretrain} (d)). 
Normalized tree edit-distance (nTED) has been introduced by Hwang et al.~\cite{nted} as a metric for comparing document parses. However, nTED is not permutation invariant, which is undesirable due to the lack of a canonical read order for forms. We introduce a modified metric, Greedily-Aligned nTED or GAnTED, which is more robust to permutation. GAnTED is discussed in detail in the Supplementary Materials. We compute GAnTED for FUDGE~\cite{fudge} by running Tesseract\footnote{\url{https://github.com/tesseract-ocr/tesseract}} on the bounding boxes it predicts and using the class and relationship predictions to build the JSON output. 

We also compare using standard 
F-measure for entity detection and relationship detection. We do this by aligning \modelName{}'s predicted strings to the GT strings. This means our results are dependant on the text recognition of \modelName{}.  This is in contrast to other models that use the GT word boxes for tokens and need only identify the correct box(es) rather than produce the correct text. Thus we end up below what prior methods achieve. Our results on both the FUNSD and NAF datasets are presented in Tables \ref{tab:funsd} and \ref{tab:naf} respectively. On the NAF dataset, no models rely on external recognition models.


The visual domain of NAF is very different from modern documents, meaning two-stage methods require a specialized recognition model. We compare Dessurt's recognition ability to a CNN-LSTM \cite{curtis} trained on the NAF dataset in the Supplementary Materials. We also report results pre-training Dessurt on images taken from the U.S.A. 1940 Census (visually similar to NAF data) in Table~\ref{tab:naf}. Details for this pre-training are in the Supplementary Materials.

\begin{table}[t]
\centering
\caption{Results on FUNSD dataset}\label{tab:funsd}
\begin{tabular}{l|l|l|l|l|l}
                               & GT OCR used              & \# params & Entity Fm & Rel Fm & GAnTED \\ \hline
LayoutLM\textsubscript{BASE}~\cite{layoutlm}& boxes + text & -         & 78.7      & 42.8~\cite{bros2021}   & -      \\
BROS\textsubscript{BASE}~\cite{bros2021}               & boxes + text & 138M + OCR      & 83.1      & \textbf{71.5}   & -      \\
LayoutLMv2\textsubscript{BASE}~\cite{layoutlmv2} & boxes + text & 200M + OCR      & 82.8      & -      & -      \\
DocFormer\textsubscript{BASE}~\cite{docformer} & boxes + text & 183M + OCR      & \textbf{83.3}      & -      & -      \\
Word-FUDGE~\cite{fudge}        & boxes                 & 17M + OCR       & 72.2      & 62.6   &    -    \\
FUDGE~\cite{fudge} (+Tesseract)             & none                       & 17M (+OCR)       & 66.5      & 56.6   &      34.8  \\
\modelName{} (ours)            & none                       & 127M      & 65.0      & 42.3   &  \textbf{23.4}     
\end{tabular}

\end{table}

\begin{table}[t]
\centering
\caption{Results on NAF dataset}\label{tab:naf}
\begin{tabular}{l|l|l|l|l}
                          & \# params & Line Fm & Rel Fm & GAnTED \\ \hline
Davis et al.~\cite{davis} & 1.8M     & \textbf{73.8}    & 49.6   & -      \\
FUDGE~\cite{fudge}        & 17M       & 73.7    & \textbf{57.3}   & -       \\
\modelName{} (ours)       & 127M      & 49.3    & 29.4   &  42.5 \\
\modelName{} w/ census pretraining   & 127M      & 50.2    & 30.3   &  \textbf{38.8}     
\end{tabular}

\end{table}

\subsection{IAM Database}

\begin{table}[t]
\centering
\caption{Results on IAM page/paragraph recognition}\label{tab:recog}
\begin{tabular}{l|r|l|l}
                                                  & \# params & CER          & WER           \\ \hline
Bluche~\cite{bluche2016joint}                     & \multicolumn{1}{|c|}{-} & 7.9          & 24.6          \\
Chung and Delteil~\cite{chung2019computationally} & \multicolumn{1}{|c|}{-}  & 8.5          &  \multicolumn{1}{|c}{-}            \\
Start, Follow, Read~\cite{wigington2018start}     & \multicolumn{1}{|c|}{-}   & 6.4          & 23.2          \\
OrigamiNet~\cite{yousef2020origaminet}            & 16.4M     & 4.7          & \multicolumn{1}{|c}{-}              \\
Vertical Attention Network~\cite{coquenet2022end} & 2.7M      & \textbf{4.5} & 14.6          \\
\modelName{}  (ours)                              & 127M      & 4.8          & \textbf{10.2}
\end{tabular}
\end{table}

There have been several specialized approaches for doing full-page handwriting recognition, where line segmentation is done implicitly or explicitly. \modelName{} is trained to do full-page recognition during its pre-training. We compare it to other full-page recognition models in Table~\ref{tab:recog}. The metrics used are character error rate (CER) and word error rate (WER) across an entire page (or paragraph; the IAM dataset has one paragraph per page). \modelName{} performs quite favorably compared to these specialized approaches and even achieves the lowest WER. 
We note that our pre-training includes synthetic handwriting derived from the IAM training set,  so \modelName{} is uniquely suited to solve this task on the IAM dataset. The fact that \modelName{}'s WER is relatively better than its CER is unusual and is likely a result of the word-part token prediction (other models use character prediction) and the language modeling capabilities learned in pre-training. We note that the number of parameters in \modelName{} is one or two orders of magnitude higher than the other models. 

\begin{table}[t]
\centering
\caption{Results on IAM NER. Reported in macro F-measure}\label{tab:ner}
\begin{tabular}{l|l|l|l|l}
 \multicolumn{1}{r|}{\begin{tabular}[c]{@{}r@{}}Split\\ Task\end{tabular}} &
  \begin{tabular}[c]{@{}l@{}}RWTH\\ 6 classes\end{tabular} &
  \begin{tabular}[c]{@{}l@{}}Custom\\ 6 classes\end{tabular} &
  \begin{tabular}[c]{@{}l@{}}RWTH\\  18 classes\end{tabular} &
  \begin{tabular}[c]{@{}l@{}}Custom\\ 18 classes\end{tabular} \\ \hline
Toledo et al.~\cite{toledo}                   & 34.0 & 37.4 & 14.9 & 18.0 \\
Rowtula et al.~\cite{rowtula}                  & 47.4 & 54.6 & 32.3 & 30.3 \\
Tüselmann et al.~\cite{tuselmann}                         & \textbf{70.7} & \textbf{76.4} & \textbf{52.0} & \textbf{53.6} \\
\modelName{}  (ours)            & 62.0 & 71.5 & 40.4 &  48.5\\
\modelName{} w/ IAM pretraining & 59.5 &71.1  & 39.5 &  45.3   
\end{tabular}

\end{table}

We also evaluate using the IAM NER task introduced by
Tüselmann et al.~\cite{tuselmann} as part of a set of named entity recognition problems for handwriting datasets. 
Tüselmann et al. use a two-stage approach constructed specifically for this problem. They use a word level handwriting recognition model, with its outputs fed to a RoBERTa-based NER model (which sees the whole document). 
We fine-tune \modelName{} on both line level NER and document level NER. In both cases \modelName{} sees the entire handwriting image 
but has the lines it is supposed process highlighted. It performs transcription along with the classification with two tasks: (1) first reading a word, and then predicting its class, and (2) the reverse with class predicted first. This ensures  we know which word \modelName{} is predicting a class for. 
We randomly replace words in the teacher-forcing with close edit-distance words to decrease reliance on the recognition output. Additionally, we apply warp grid augmentation~\cite{curtis} on the lines of the document. We also experimented with adding recognition on IAM words to the pre-training (more details in Supplementary Materials).

Our results for IAM NER are presented in Table~\ref{tab:ner}. While \modelName{} is moderately successful, it falls short of the customized two-stage approach presented by Tüselmann et al.  They report that the CER of the HWR model they use is 6.8, which is the same CER as \modelName{}. We assume this indicates that (unsurprisingly) RoBERTa is a stronger language model than \modelName{} and is responsible for this superior performance. 

\subsection{Ablation}

We performed an ablation study of the different sources used in our model's pretraining as well as some of the architectural choices (Table~\ref{tab:abl}). We begin the data ablation with only the IIT-CDIP~\cite{iit} dataset (I). We then incrementally add the synthetic Wikipedia (W), synthetic handwriting (H), synthetic forms (F), and distillation from BART (D). We ablate out the the predicted spatial mask used in pretraining,  and change the Swin window size from 12 to 7. We also ablate the 2-way cross attention by instead only having the query and response tokens attend to the visual tokens without the visual tokens attending to the query tokens. This is very similar to Donut, which lacks 2-way cross attention.


As can be seen each pre-training data source adds something to the model. The synthetic handwriting and synthetic forms are aimed at particular downstream tasks (IAM NER and form understanding respectively), but we note that their inclusion generally helps other tasks as well. Only the distillation appears selectively helpful and may not contribute significantly.
In general, the ablated model components are helpful to the full model, but not necessary. The results with the RVL-CDIP dataset shows that the data a model is pre-trained with appears to be relatively irrelevant to its performance.

\begin{table}[t]
\centering
\caption{Ablation results. The top four rows show the pre-training ablation with I=IIT-CDIP dataset, W=synthetic Wikipedia dataset, H=synthetic handwriting dataset, F=synthetic form dataset, D=distillation from BART. The lower three rows show ablations to the model: removing supervision with output mask, removigin supervision with output mask and reducing Swin window size to 7, removing cross attention from image to question tokens. Results for DocVQA are evalutated using the validation set. ``PT IAM'' indicates IAM data added to last 200k iters of pre-training}\label{tab:abl}
\begin{tabular}{r|c|c|c|c|c|c|c|c}
\multicolumn{1}{l|}{} &


  \multirow{3}{*}{\begin{tabular}[c]{@{}l@{}}DocVQA\\ (valid)\\ ANLS\end{tabular}} &
  \multicolumn{2}{c|}{IAM NER} &
  \multicolumn{2}{c|}{FUNSD} &
  \multicolumn{2}{c|}{NAF} &
  \multirow{3}{*}{\begin{tabular}[c]{@{}l@{}}RVL \\ CDIP\\ acc.\end{tabular}} \\ 
  
  &   &  \multicolumn{2}{c|}{Macro Fm}  &  Entity &  Rel &  Entity &  Rel &  \\ 
  &   &  &   IAM PT &  Fm &  Fm &  Fm &  Fm &  \\ \hline
  
  \multicolumn{1}{l|}{Max iterations} & 500k & 200k & 200k & \multicolumn{2}{c}{34k}    & \multicolumn{2}{|c|}{300k}    & 500k \\ \hline
I                      & 44.0 & 42.3          &  43.4         & 19.7          & 10.2          & 28.7          & 12.6          & 89.0          \\
W+I                    & 43.2 & 45.2          & 49.0          & 29.5          & 16.0          & 31.0          & 13.7          & 89.1 \\
H+W+I                  & 44.4 & 50.1          & 49.7          & 29.3          & 16.5          & 31.6          & 14.9          & 88.9          \\
F+H+W+I                & \textbf{46.5} & 47.6          & 50.0          & 44.8          & 28.2          & \textbf{36.5}          & \textbf{17.6}          &       \textbf{89.5}        \\
D+F+H+W            & 43.1 & \textbf{52.7} & \textbf{53.3} & 39.4 & 22.0 & 31.5 & 14.3          & 88.5          \\
All=D+F+H+W+I            & 45.5 & 50.4 & 52.5 & \textbf{47.8} & \textbf{29.5} & 34.6 & 15.3          & 89.0          \\
All, no mask loss      & 44.9 & 45.7          & 49.7          & 47.3          & 26.2          & 33.2          & 15.1          & 88.3          \\
All, no mask loss w=7  & 44.4 & 44.8          & 51.3          & 45.9          & 28.6          & 31.8          & 15.3          & 88.6          \\
All, 1-way cross attn. & 44.9 & 42.9          & 46.9          & 41.0          & 25.2          & 33.7          & 15.9 & 88.8         
\end{tabular}

\end{table}

\section{Conclusion}
We have introduced Dessurt, an end-to-end architecture for solving a wide variety of document problems. Dessurt performs recognition within its single pass, removing reliance on an external recognition model, which most document understanding approaches require, making it a much simpler method. Because Dessurt uses arbitrary text as its output, it is also more flexible in the range of problems it can solve. We evaluate Dessurt on a wider range of tasks than any previous single method has done and show results ranging from promising to state-of-the-art.

\clearpage
%
%
\bibliographystyle{splncs04}
\bibliography{egbib}
\end{document}


\pagestyle{headings}
\mainmatter
\def\ECCVSubNumber{6561}  

\title{Supplementary Materials for \\ End-to-end Document Recognition and Understanding with \modelName{}} 

\titlerunning{Supplementary Materials for \modelName{}}
%
\author{Brian Davis\inst{1} \and
Bryan Morse\inst{1} \and
Bryan Price\inst{2} \and 
Chris Tensmeyer\inst{2} \and
Curtis Wigington\inst{2} \and
Vlad Morariu\inst{2}}
\authorrunning{B. Davis et al. (Supplementary Materials)}
%
\institute{Brigham Young University, Provo UT, USA 
\email{\{briandavis,morse\}@byu.edu}
\and
Adobe Research, USA
\email{\{bprice,tensmeye,wigingto,morariu\}@adobe.com}}

\maketitle

In these Supplementary Materials we provide details on the model, training, and evaluation of Dessurt. There also additional examples of the synthetic data.
The contents are as follows:
\begin{enumerate}
    \item Model Details: Specific details on the Dessurt architecture
    \item Pre-training Procedure
     \begin{enumerate}[label*=\arabic*.]
     \item IIT-CDIP Dataset: Details on our OCR process and all tasks used in pre-training
     \item Synthetic Wikipedia: Details on the paragraph generation and the font database
     \item Synthetic Handwriting: Details on handwiting generation
     \item Synthetic Forms: Details on GPT-2 generation, form generation (layout), JSON parse format, amd pre-training tasks
     \item Selection of ``Easy'' Fonts for Distillation
     \item Pre-training Curriculum Details
     \end{enumerate}
    \item Data Augmentation
    \item GAnTED: A full description of the modified nTED metric we introduce
    \item Experiment Details: Additional details for each evaluated task
\end{enumerate}




\section{Model Details}

\begin{figure}[t]
\centering
\includegraphics[width=0.99\textwidth]{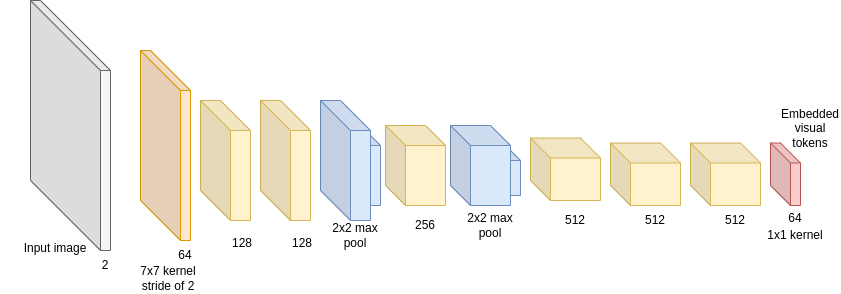}
\caption{Architecture of the visual encoder. Each convolution layer (orange, yellow, red) except the last is followed by Group Norm~\cite{groupnorm}, dropout, and ReLU. The last convolution is followed by Layer Norm. All yellow convolution layers have 3x3 kernels}
\label{fig:visualencoder}
\end{figure}

 The down-sampling/tokenization CNN used by Dessurt was inspired by the CNN component of a CNN-LSTM~\cite{curtis}. We originally pre-trained the CNN as part of a line recognition OCR model, but found that this did not improve training (it could learn just as well from scratch with small images).  The CNN down-samples the input image by a factor of 8, meaning the input visual tokens of Dessurt are $144 \times 96$ = 13,824 total.
 This CNN has 7 convolution layers detailed in Figure~\ref{fig:visualencoder}.
 We found having an aggressive down-sampling on the computationally light CNN (as opposed to Swin layers) was vital for being able to fit our model in memory when running on large images.
 
 Dessurt has 8 full cross attention layers layers and 2 more which only update the textual tokens (using the last visual tokens). We down-sample the visual tokens (as Swin~\cite{swin} does) after the first 4 layers. The initial width of the visual tokens is 128 and the becomes 256 after the down-sampling. We note this is quite small and was necessary due to the large image size we process. The width of the textual tokens is 768. The reverse-bottleneck of the fully connected layers for the textual tokens goes up to 3072.
 The model processes a maximum of 20 query tokens and 800 response tokens. The response must be long for form parsing, where the entire document is predicted.

 Swin~\cite{swin} doesn't use 2D position embedding, but instead relies on relative position attention bias. We include 2D spatial embedding because the textual tokens are attending to the visual tokens and may need location information.

\section{Pre-training Procedure}

\subsection{IIT-CDIP Dataset}

There isn't a standard OCR for the IIT-CDIP~\cite{iit} used by researchers and it hasn't been investigated what impact this might make in pre-training.. 
We processed the dataset using Tesseract\footnote{\url{https://github.com/tesseract-ocr/tesseract}}, an open source OCR engine. Tesseract makes many errors and doesn't capture the layout of the documents very well. We perform some post processing including rotating the image upright and attempting to extract the block or paragraph structure by doing layout analysis using Publaynet~\cite{publaynet} and PrimaNet~\cite{antonacopoulos2009realistic} models available on LayoutParser~\cite{layoutparser}. We will make try to make our OCR results available for other researcher as the dataset is quite large and this is a very long process.

To check rotation, we examine the average confidence score $C_{mean}$ returned by Tesseract as well as the average width-to-height ratio for the returned word boxes $\frac{width_i}{height_i} = R_i$. If $C_{mean} > 80$ and $R_{mean} > 1$ we assume the rotation is correct. If not we then run Tesseract on $90^{\circ}$, $270^{\circ}$, and $180^{\circ}$ rotations of the image. If any passes the before-mentioned threshold, we accept it as the correct rotation. If none do, we compare the product score $C_{mean}R_{mean}$ of the four rotations and accept the one with the largest product score above 55. If none have a product score above 55, the image is removed from the dataset. This removes images without words and images Tesseract particularly struggled with. When we accept a rotation, we not only use that OCR result, but also use that rotated image in the dataset.

We discard Tesseracts block and paragraph groupings.
We then run both the Publaynet and PrimaNet models on the image and append the returned bounding boxes. We remove ``super'' layout boxes that are superfilous by removing any bounding box covering more than 90\% of the text boxes when most text lines overlap with multiple layout bounding boxes.
We then go though each text line and assign it to a layout bounding box based on how many lines also overlap with that bounding box and how full the layout bounding box is with text lines. We then collapse the layout bounding boxes to the text lines assigned to them. We then find highly overlapping layout bounding boxes and merge them together. These remaining layout bounding boxes, and any text lines which were not assigned a layout bounding box, are the final blocks used as our layout annotations for the IIT-CDIP data.

Because some tasks are reliant on the block or paragraph structure of a document, and sometimes the extracted block structure is poor, we look at the area of each block covered by its words to heuristically decide if the block structure was accurately extracted or not.
For each block we compute height to width ratio (tall is good as it probably has multiple lines) and how much of the block is covered by its text lines (more is better as its dense text). These get averaged, with each block getting weight equal to the number of text lines in it. If this above a threshold, we accept the block structure as good for layout based tasks.

If not stated otherwise, the model is supervised to predict a pixel mask for whatever text it reads (outputs). Wherever text is removed from the document image, has a -1 on the input mask (also called highlight mask).

\begin{figure}[pht]
\centering
\includegraphics[width=0.99\textwidth]{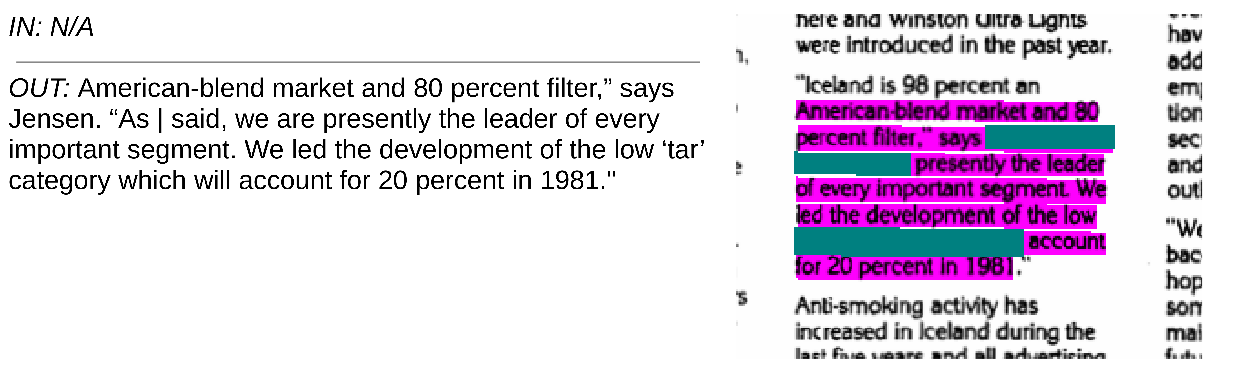}
\caption{Example Text Infilling task: Magenta is highlight, turquoise is deleted text.}
\label{fig:textinfilling}
\end{figure}
\begin{figure}[pht]
\centering
\includegraphics[width=0.99\textwidth]{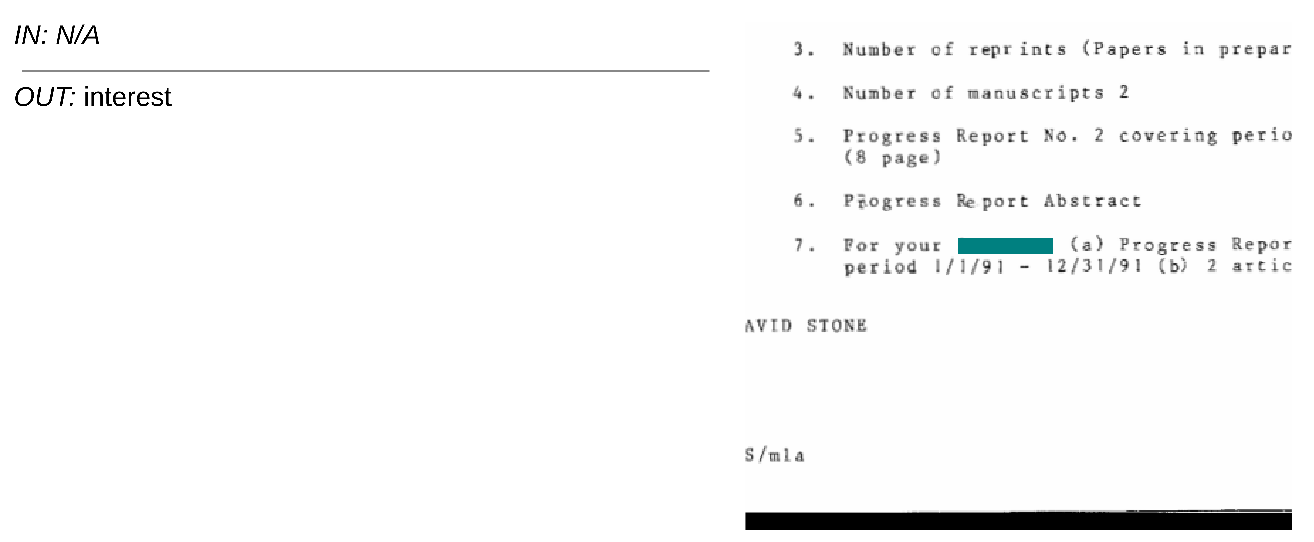}
\caption{Example Word Infilling task: Turquoise is deleted word.}
\label{fig:wordinfilling}
\end{figure}
\begin{figure}[pht]
\centering
\includegraphics[width=0.99\textwidth]{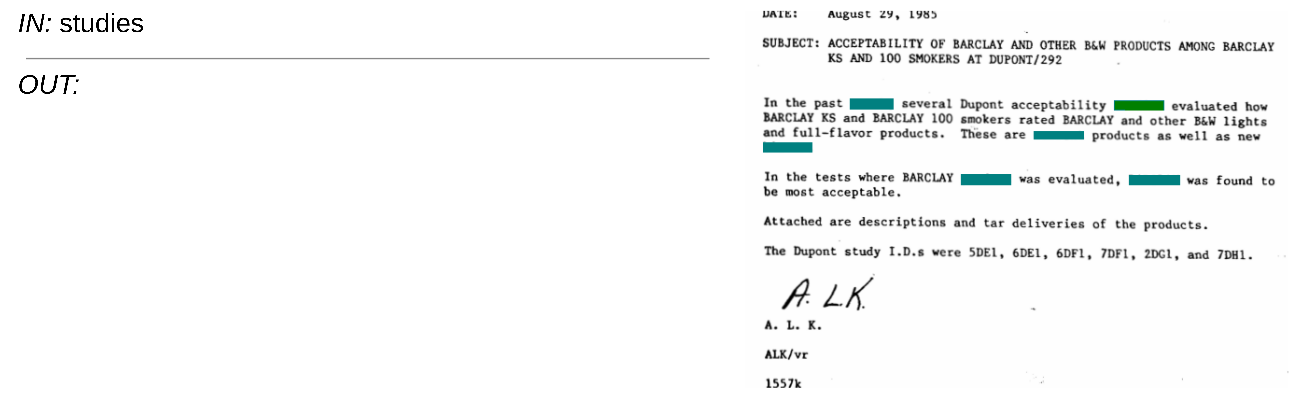}
\caption{Example Place Word task: Turquoise is deleted words, green is GT output mask.}
\label{fig:place}
\end{figure}
\begin{figure}[pht]
\centering
\includegraphics[width=0.99\textwidth]{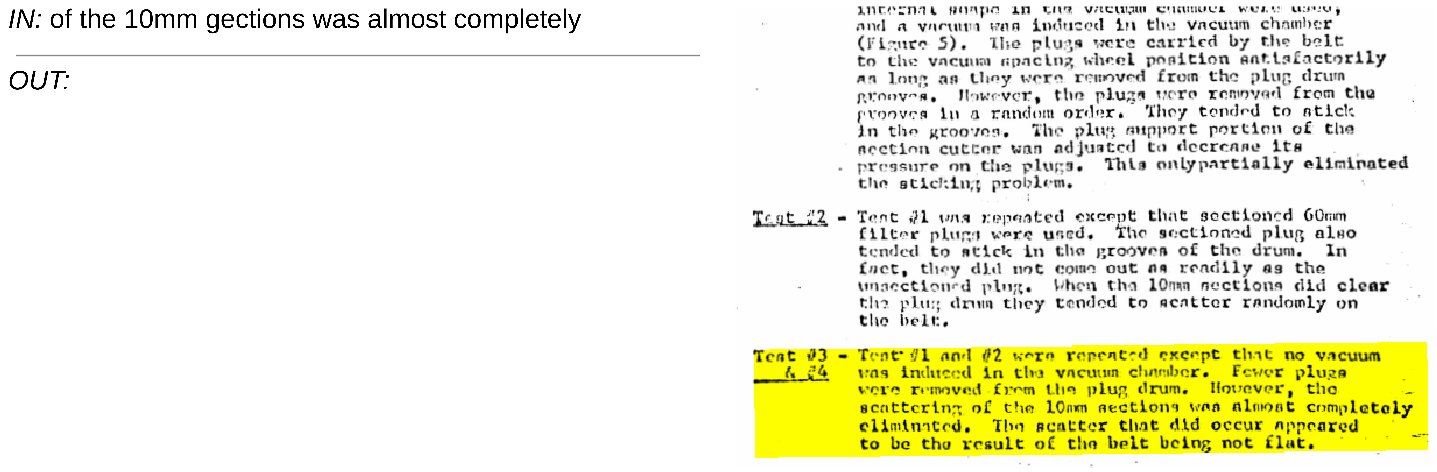}
\caption{Example Highlight Block task: Yellow is GT output mask.}
\label{fig:block}
\end{figure}
\begin{figure}[pht]
\centering
\includegraphics[width=0.99\textwidth]{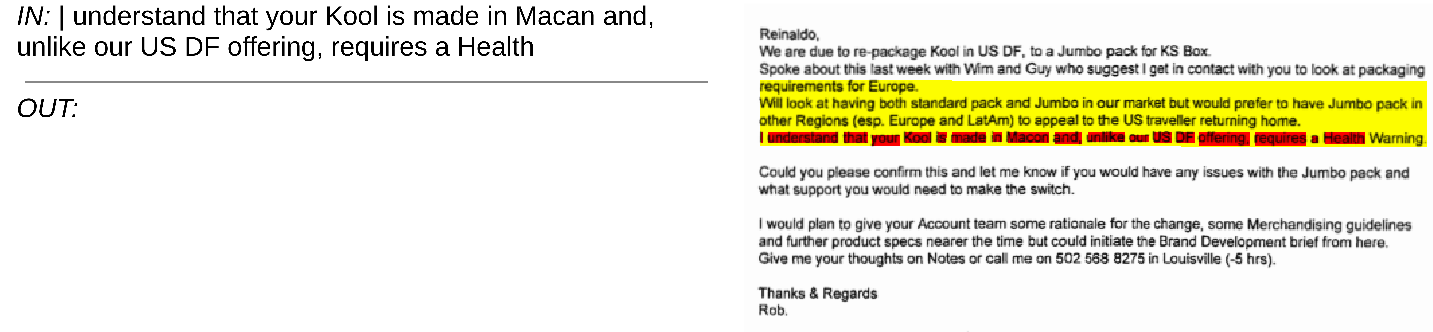}
\caption{Example Highlight Block task (with input highlight): Red is highlight input, Yellow is GT output mask. Notice this has a block segmentation error.}
\label{fig:block0}
\end{figure}
\begin{figure}[pht]
\centering
\includegraphics[width=0.99\textwidth]{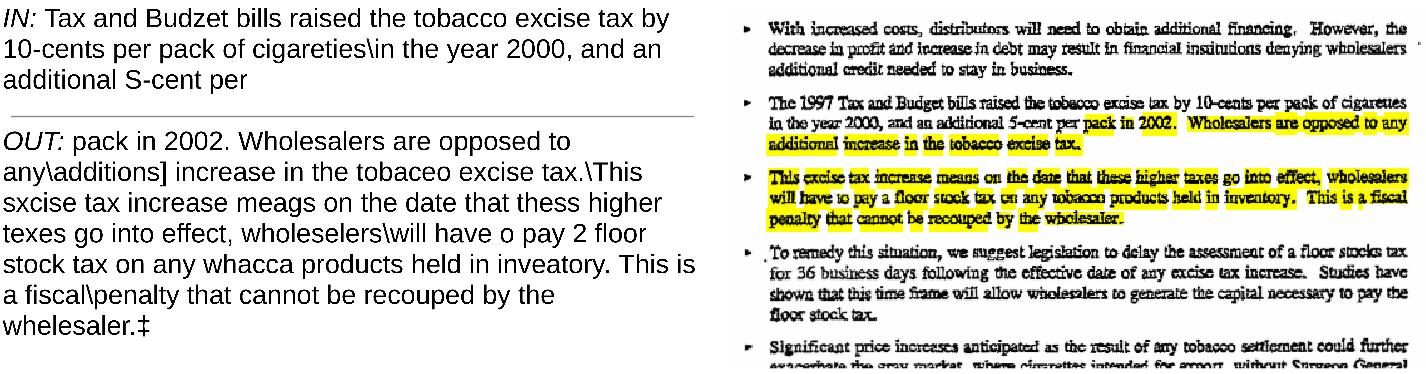}
\caption{Example Read On task: Yellow is GT output mask.}
\label{fig:readon}
\end{figure}
\begin{figure}[pht]
\centering
\includegraphics[width=0.99\textwidth]{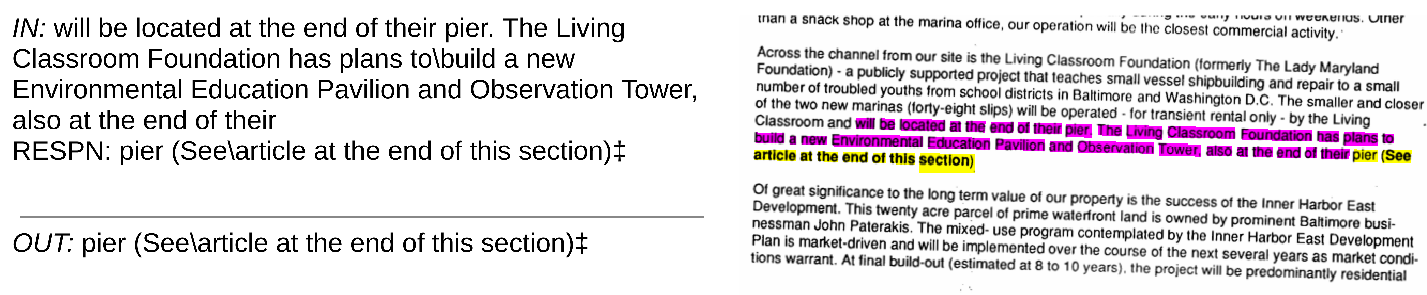}
\caption{Example Read On task (with input highlight): Magenta is highlight input, Yellow is GT output mask.}
\label{fig:readon0}
\end{figure}
\begin{figure}[pht]
\centering
\includegraphics[width=0.99\textwidth]{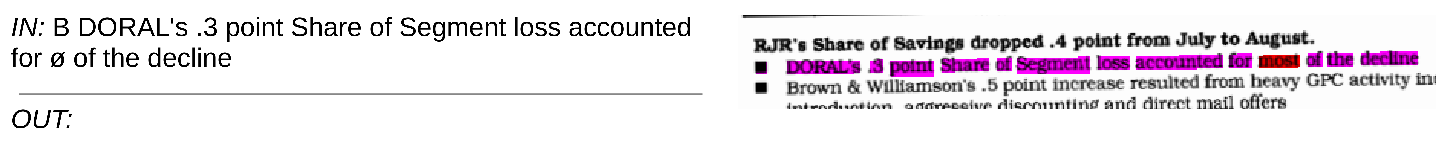}
\caption{Example Get Blanked task: ``$\emptyset$'' is blank character. Magenta and red are highlight input. Red is GT output mask.}
\label{fig:blanked}
\end{figure}
\begin{figure}[pht]
\centering
\includegraphics[width=0.99\textwidth]{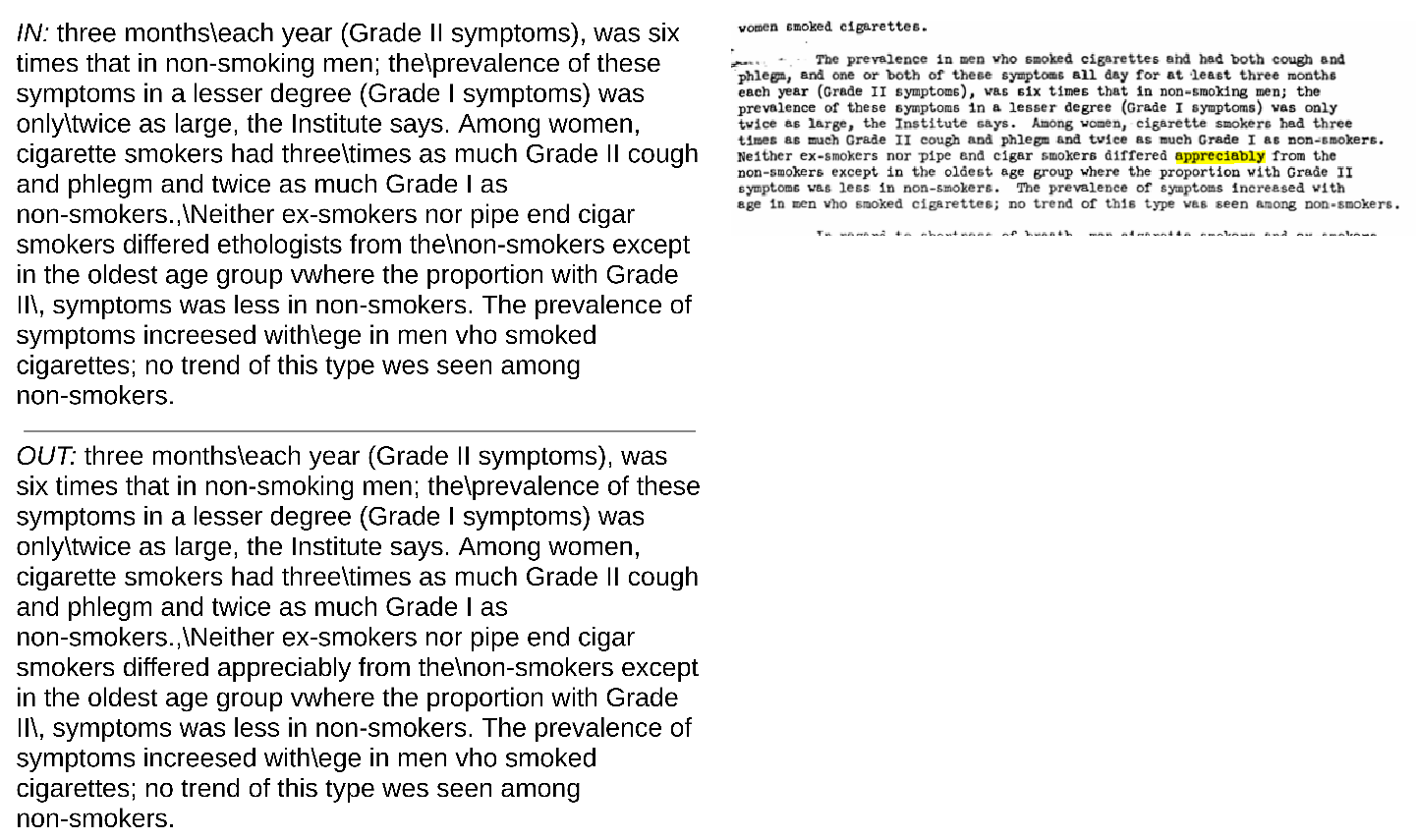}
\caption{Example Re-read Replaced task: Yellow is GT output mask.}
\label{fig:reread}
\end{figure}
\begin{figure}[pht]
\centering
\includegraphics[width=0.99\textwidth]{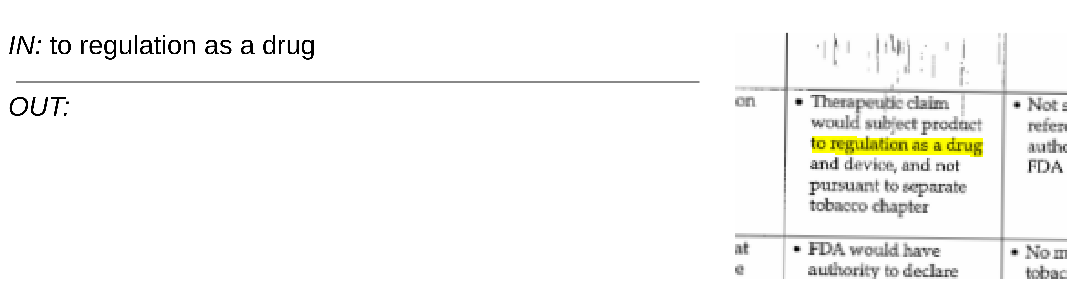}
\caption{Example Highlight task: Yellow is GT output mask.}
\label{fig:highlight}
\end{figure}
\begin{figure}[pht]
\centering
\includegraphics[width=0.99\textwidth]{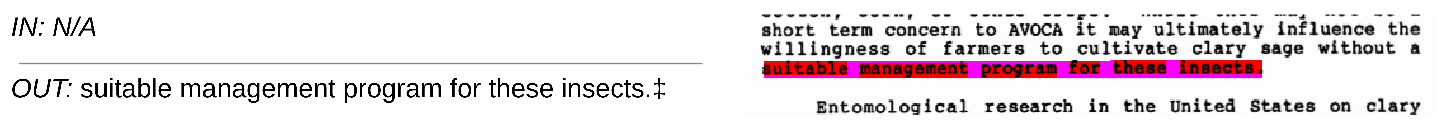}
\caption{Example Read Highlight Text task: Magenta and red are highlight input. Red is GT output mask.}
\label{fig:read}
\end{figure}

We now list all the tasks for pre-training with the IIT-CDIP dataset, and their frequency. One can note in the provided Figures many errors resulting from our OCR and layout analysis steps.  Tasks with ``*'' require good block annotation.
\begin{itemize}
    \item 66.1\% Text Infilling (Fig.~\ref{fig:textinfilling}): This is a MLM task inspired by the text infilling used to train BART~\cite{bart}; however instead of replacing the removed text with a blank token, we delete them from the image, replacing them with white. The area is marked so the model knows something was removed and the entire text block is highlighted. The model then must predict the text of the entire block, filling in the blanked regions. This is easier than the infilling task used by BART for two reasons: (1) the length of what should be filled in can be approximated by the physical blank-space, and (2) we do not allow a blank area of 0 tokens (inserted blank token).
    \item 16.5\% Word Infilling (Fig.~\ref{fig:wordinfilling}): This is a potentially more difficult MLM task. A single word in the document is removed and the model must predict that word. In the above task the model is forced to place the text in context by generating the entire block. For this task it predicts the word in isolation, and thus must capture language context in its hidden states somewhere.
    \item 4.1\% Place Word (Fig.~\ref{fig:place}): A different flavor of MLM. Several words, of roughly the same length, are removed from the document image. The query contains one of the removed words.  The model must predict a pixel mask at the location(s) the given word occurs.
    \item 4.1\% Highlight Block* (Figs.~\ref{fig:block} and ~\ref{fig:block0}): The query contains a small snippet of text (and randomly the text is highlighted in the input) and the model must predict a pixel mask covering all the words in the block the text belongs to. This is intended to teach document layout. 
    \item 3.7\% Read On* (Figs.~\ref{fig:readon} and~\ref{fig:readon0}): The query contains a short text snippet (and randomly the text is highlighted in the input) and the model is to read starting after that text, following newlines, until the end of the block. This teaches text recognition from both a finding and reading standpoint.
    \item 2.1\% Get Blanked (Fig.~\ref{fig:blanked}): The query has a snippet of text, but one word is replaced by a blank token. The model must read the word that fits in the blank token. It randomly has the text snippet highlighted or not.
    \item 2.1\% Re-read Replaced (Fig.~\ref{fig:reread}): The model is given a snippet of text, but one word is replaced by a random word of the same length. The model then must read the text using the correct word. It randomly has the text snippet highlighted or not.
    \item 0.4\% Highlight Text (Fig.~\ref{fig:highlight}): The query has a snippet of text and the model predicts a mask for it.
    \item 0.4\% Read Highlight (Fig.~\ref{fig:read}): A text line is highlighted for the model to read.
    \item 0.2\% Read Line Above: The query has a snippet of text and the model must read the text line above it.
    \item 0.2\% Read Line Below: The query has a snippet of text and the model must read the text line below it.
\end{itemize}

\begin{figure}[pht]
\centering
\includegraphics[width=0.99\textwidth]{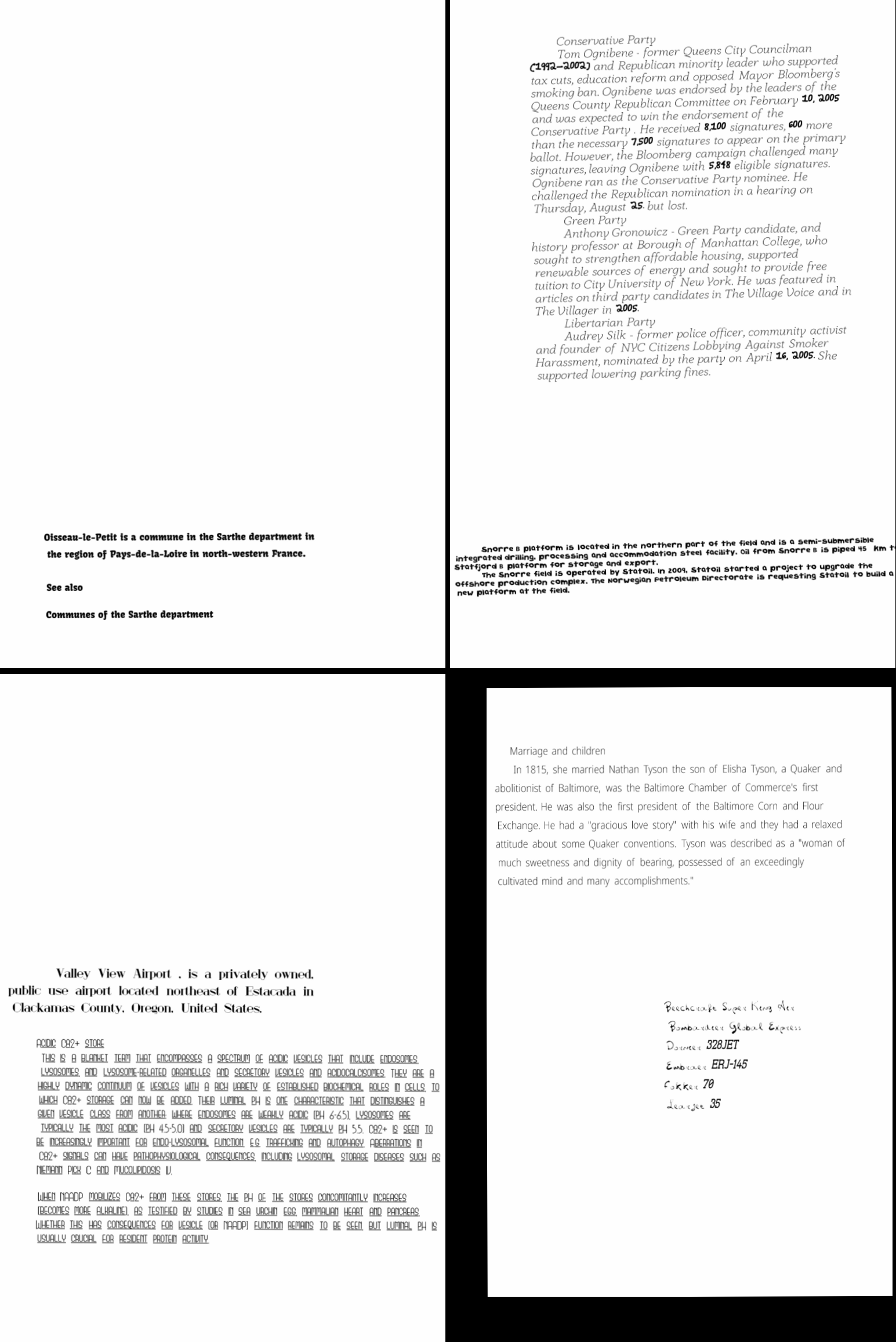}
\caption{Examples of synthetic Wikipedia documents}
\label{fig:synthwiki}
\end{figure}

\subsection{Synthetic Wikipedia}

We first detail the paragraph generation method used in creating the synthetic documents. We then discuss the collected font database in more detail. We also include additional synthetic document examples in Fig.~\ref{fig:synthwiki}.

\subsubsection{Details on paragraph generation}

The document begins as a blank image the same size as our model input.

The column width is sampled in the range of the whole image width to 1/5 of the image. The text height is from the range of 8 to 32 pixels. We note that at 8 pixels, many fonts are illegible. When rendering text, we estimate the maximum height of that font by generating a placeholder string with ascenders and descenders, and the scale to resize this placeholder string to the selected text height is the scale used to resize the actually rendered text. We predict spacing based on an approximated Em at 1.6 times the text height\footnote{\url{https://en.wikipedia.org/wiki/Em\_(typography)}}, and then the minimum and maximum (horizontal) space as 0.2 to 0.5 times the Em\footnote{\url{https://docs.microsoft.com/en-us/typography/develop/character-design-standards/whitespace}}. The newline spacing is sampled between 1 pixel and the text height.

Each word is generated individually and then they are arranged in paragraph form, placing words in a line until the column width is reached and then wrapping onto a new line. There three different paragraph formats selected with the following probabilities: indented 80\%, no indent 18\%, inverse indent 2\%. On an intended paragraph format we select and indent length from 0.3 to 6.0 times the Em and each first line of a paragraph is indented accordingly. For no indent, extra space is added at a newline, randomly from 0 to the selected newline height, whenever a new paragraph is starting. For inverse indent, all lines except the first are indented.
When starting a newline, we randomly add a perturbation indent, from 0 to the horizontal space width, to add noise to the process.

If the height of the rendered article exceeds the image height, we increase the column width (and resample the horizontal, newline, and indent spacing) and replace the words.

Articles are repeatedly added to the image until one cannot be placed.

\subsubsection{Font database}

The 10,566 fonts we scrape from 1001fonts.com are not curated and so some fonts are not actual text fonts (Wingdings-like). Many don't include numbers and/or punctuation and some have only upper case letters (the BART tokenization~\cite{bart} we use is case sensitive). We test fonts to automatically determine some these features and take them into account when rendering.
When we randomly select a font, if the selected font does not have numbers another font with numbers is select and is used whenever a word has a number. If the selected font has only uppercase, all GT text is converted to be uppercase.

\begin{figure}[tp]
\centering
\includegraphics[width=0.99\textwidth]{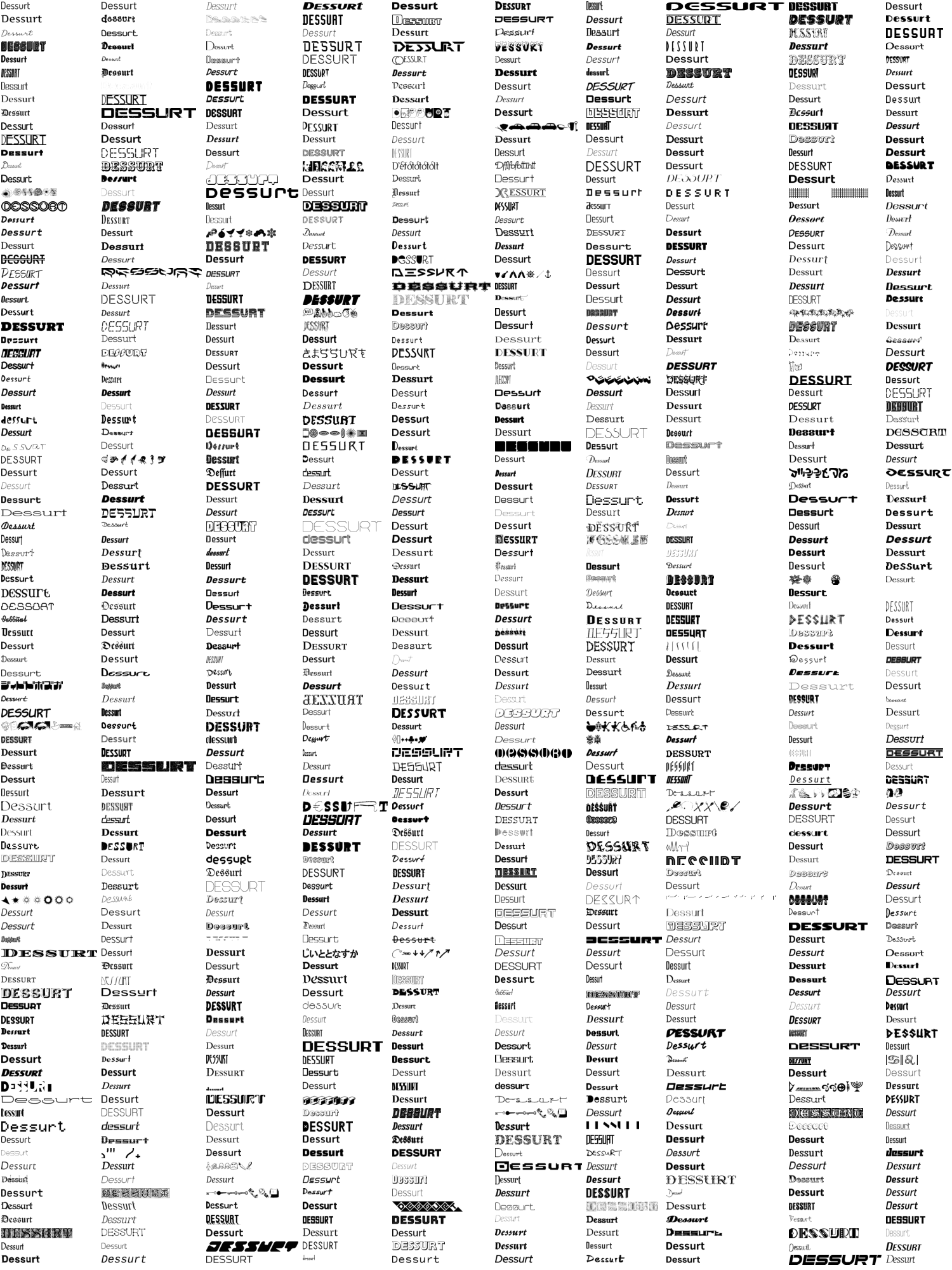}
\caption{Rendering of the word ``Dessurt'' in 949 fonts from our database at a text height of 16}
\label{fig:fonts}
\end{figure}

There are a wide variety of fonts, including handwritten and stylized fonts, but we did not track metadata when scraping, so we don't metrics on the database's distribution. However, we do render 949 fonts in Fig.~\ref{fig:fonts} as a qualitative sample. Some of the fonts are variants of others (bold or italicized).

All the code for scraping and pre-processing the fonts will be included in our released code.

\subsection{Synthetic Handwriting}
The full line handwriting synthesis method we use \cite{line_gen} to generate handwriting does not use a random style vector as input, but rather a distribution from the ones extracted from the data. We interpolate styles extracted from the IAM training set, the ``Random'' option in the generation script provided by the authors of 
\cite{line_gen}.

We note there are more realistic handwriting generation works more recently developed, but  \cite{line_gen} is the only one to generate full lines and has a convenient script in its released code for generating a dataset like this.

For the full page recognition task, in training half of the instances have the handwriting lines highlighted.

\begin{figure}[tp]
\centering
\includegraphics[width=0.99\textwidth]{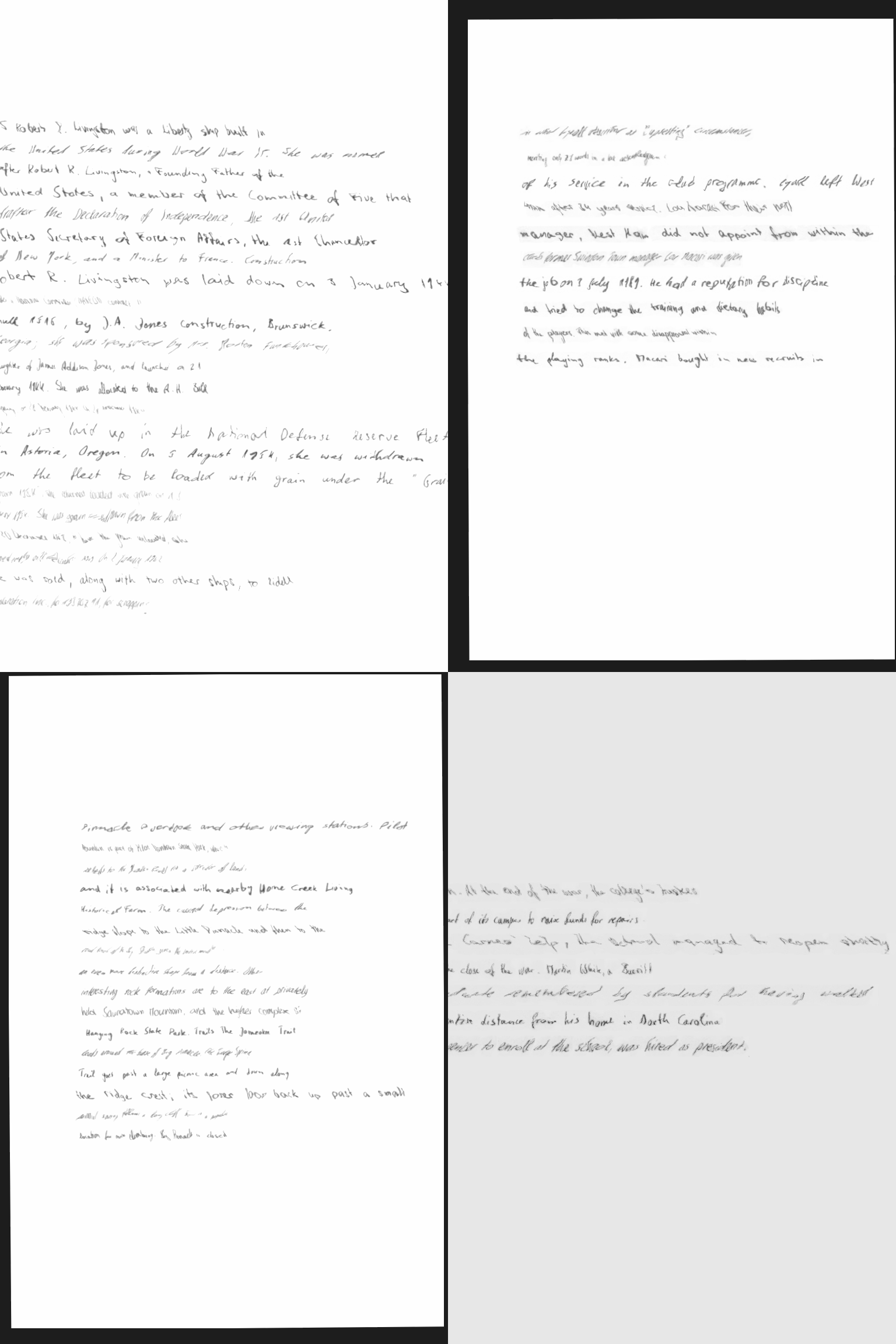}
\caption{Four rendered pages with synthetic handwriting}
\label{fig:hw}
\end{figure}

Additional examples of documents with synthetic handwriting can be seen in Figure~\ref{fig:hw}.

\subsection{Synthetic Forms}

\begin{figure}[pt]
\centering
\includegraphics[width=0.99\textwidth]{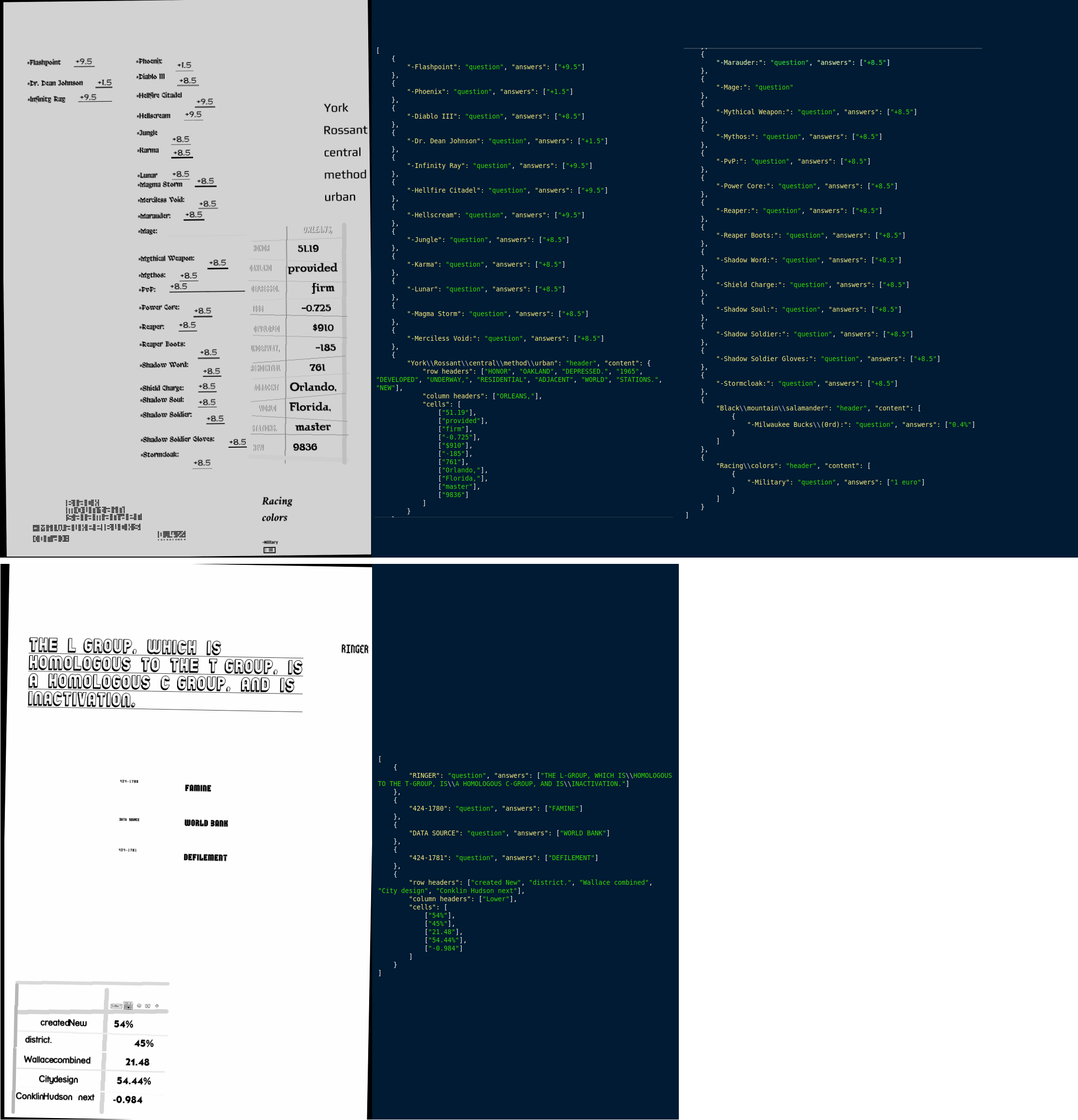}
\caption{Examples of synthetic form documents and their JSON parse}
\label{fig:synthform1}
\end{figure}

\begin{figure}[pt]
\centering
\includegraphics[width=0.99\textwidth]{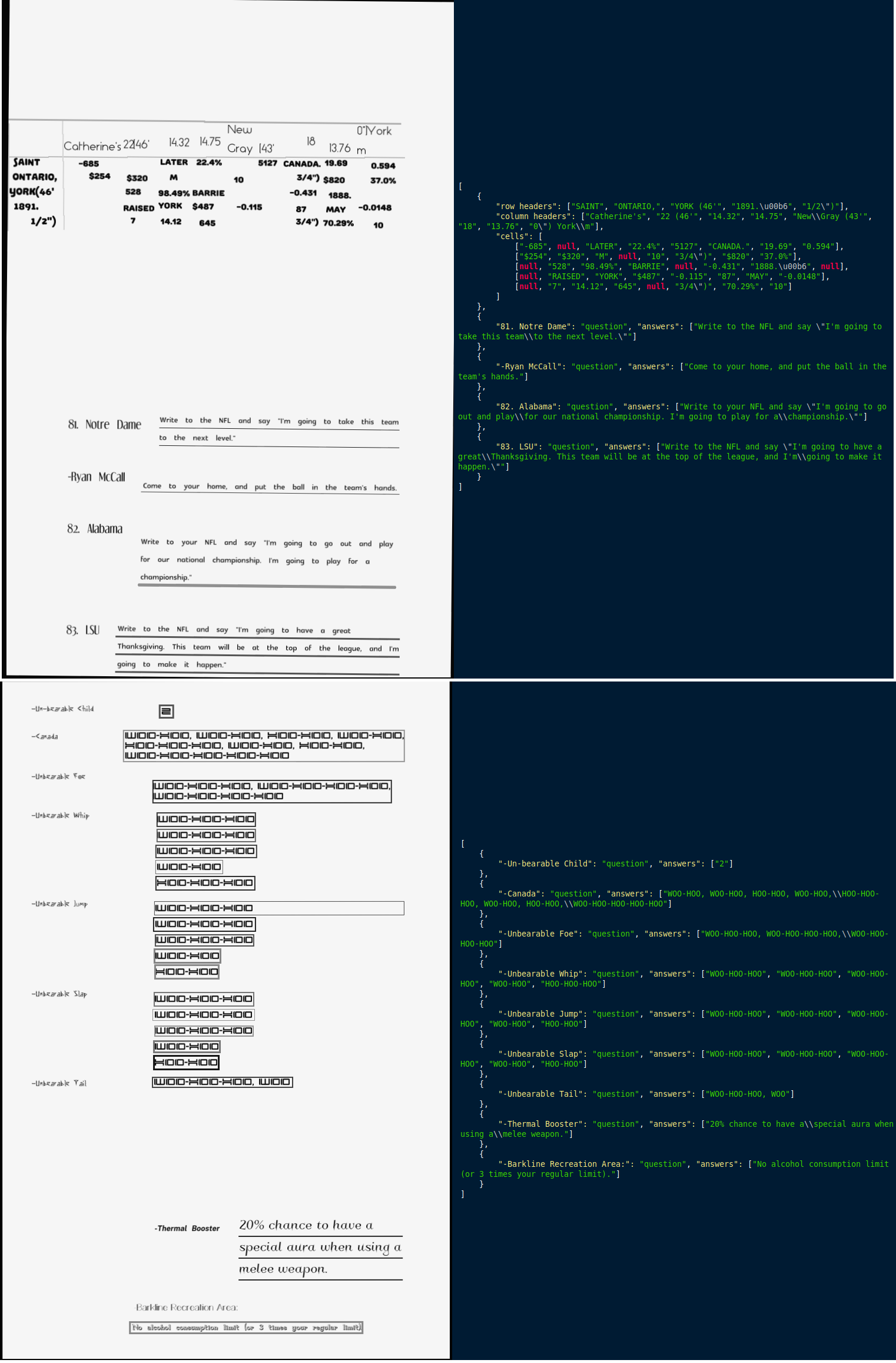}
\caption{Examples of synthetic form documents and their JSON parse. On the lower image we see GPT-2's degenerate repeated text. }
\label{fig:synthform2}
\end{figure}

Here we include the details of the generation of label-value sets with GPT-2, and how they are rendered into documents. We also include additional examples of generated images and their parse JSON in Figs.~\ref{fig:synthform1} and~\ref{fig:synthform2}.

\subsubsection{GPT-2 generation details}

The typical process for generating text with an autoregressive model is to intialize the text generation with a prompt. In our case we use both a text prompt and a ``structure'' prompt, which is an example of what format the label-value pairs should be in.
The text prompts used are:
\begin{itemize}
    \item ``This form is to be filled out.''
    \item      ``This form has been filled out.''
    \item      ``Form $X$'' where $X$ is replaced half of the time with a random letter and the other half with a letter followed by a random integer less than 10,000
    \item      ``This form is to be filled out regarding $Y$.'' where $Y$ is replaced with a random Wikipedia article title
    \item      ``This form contains information about $Y$.'' where $Y$ is replaced with a random Wikipedia article title
\end{itemize}

We draw from the pool of generated labels and label-value pairs for the structure part of the initialization for a generation run. These pools to not contain duplicate entries. Because labels with numbers can have several thousand ``near duplicates'' we limit the number of labels with numbers to be about 0.002 of the pools.
We initialize the label-value pool with 60 instances of ``Date: $X$'', where $X$ is a random date with one of 6 formats.
We initialize the label pool with: ``Name:'', ``Location:'', and ``Details:''

To generate a label-value set, we first sample a text prompt, a label-value pair, and a label and compose them as the input for GPT-2 using Huggingface's interface\footnote{\url{https://huggingface.co/gpt2}}. We use a temperature of 0.85 and generate three outputs. For each output, we parse it into label-value pairs until the parsing fails (it will frequently degenerate/stop generating a form). The parsing attempts to prevent repeated values from being added (a frequent degeneration of autoregressive models) and will also parse a comma separated list of values. List values are generated in vertical/newline separated format when creating a synthetic form. 

We generate 813,793 label-value sets, with over 7 million total label-value pairs.

We note that we accidentally split URLs into label value pairs (with the label of ``http''). We filter these out in the document creation process.

\subsubsection{Form generation details}

A synthetic form is generated by repeatedly adding a label-value set or table in an empty region of the image. After each empty region has had a failed generation, the document is complete. One empty region is the area of the document right of the rightmost content. Whenever a table or label-value set is added, a new empty region is created underneath it spanning the same horizontal space. Each label-value set is generated to fit the region it is being generated in, so this process attempts to pack the form densely.

\emph{Label-value set: \ } 
There are three possible fonts selections, for the header, labels, and values, however 30\% of the time the label font will be forced to be the same as the header font, and 50\% of the time the value font will be forced to be the same as the label. This is to make the parsing more difficult, and does reflect a frequent scenario in the FUNSD dataset~\cite{funsd}.
All labels and all values in a set will be rendered with the same respective font.

In 0.5\% of rendered label-value pairs, we replace all values with binary checkboxes. The are rendered with boxes, parentheses, or brackets (depending on what the font has) and an `X' or blank value.

A block width is randomly selected, but will be increased if the generation fails to place any label-value pairs. If the placement fails at the maximum width for the empty region, the region has a failed generation.

\begin{figure}[pt]
\centering
\includegraphics[width=0.99\textwidth]{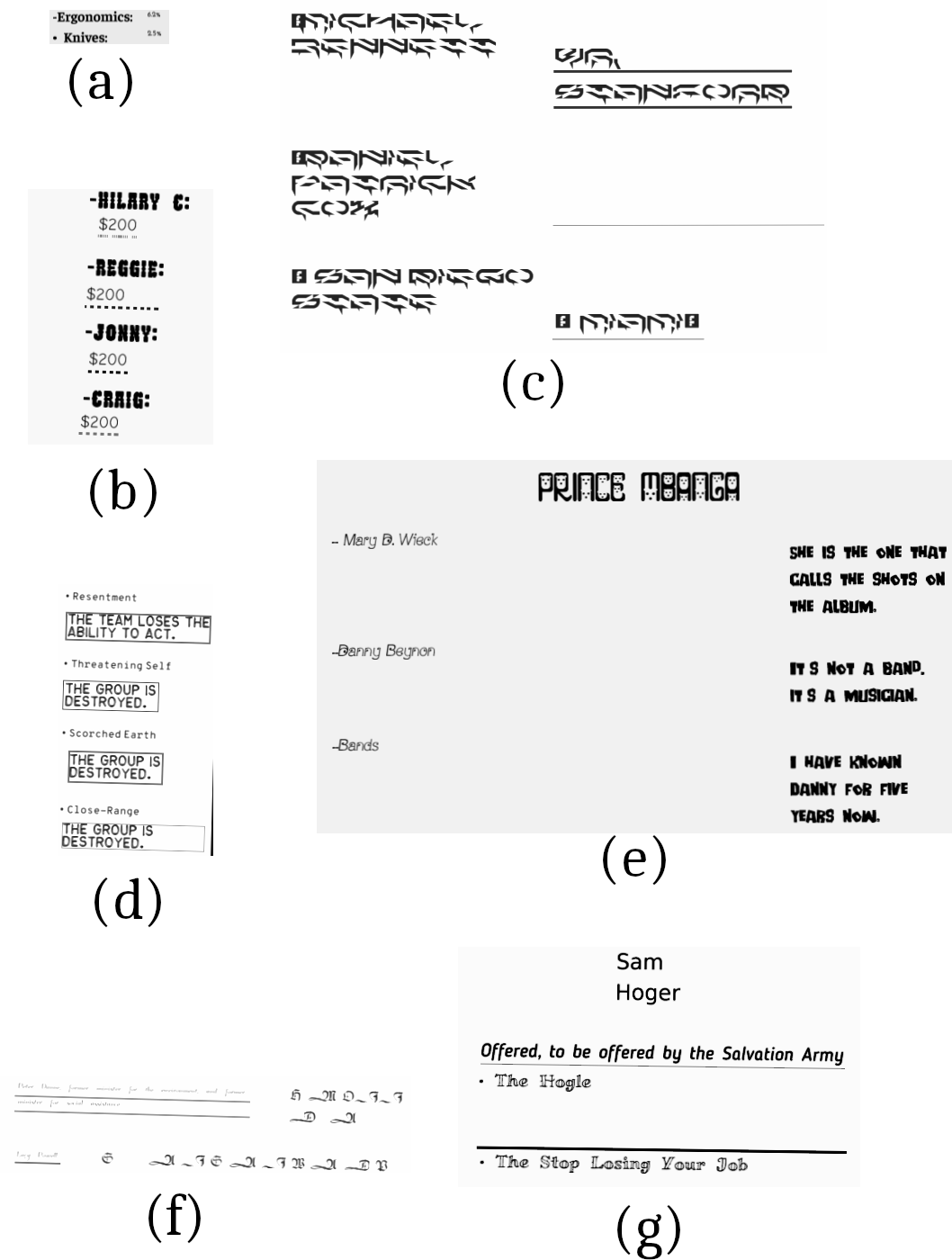}
\caption{Examples of label-value relationships. (a) Colon, (b) Colon+Dotted line, (c) Line, (d) Box, (e) To Right (with header), (f) To Left, (g) Below (with header) }
\label{fig:rel_examples}
\end{figure}

A uniformly random selection is made between 9 different relationship indicators which determine how the label-value pairs will be rendered in relation to one another.   These are the possible relationships:
\begin{itemize}
    \item Colon: A colon is added to the end of the label. See Fig.~\ref{fig:rel_examples} (a)
    \item Line: An underline is added beneath the value (or a blank area). Line thickness randomly selected per pair. See Fig.~\ref{fig:rel_examples} (c)
    \item Colon+Line: Both of the above
    \item Dotted line: A dashed or dotted underline. Frequency of dotting randomly selected.
    \item Colon+Dotted line: See Fig.~\ref{fig:rel_examples} (b)
    \item Box: The value is put in a box. Thickness of box lines is randomly selected per pair. See Fig.~\ref{fig:rel_examples} (d)
    \item Colon+Box
    \item To Right: The values will be to the right of the label with the values and labels aligned horizontally and no other cues. See Fig.~\ref{fig:rel_examples} (e)
    \item To Left: The value will be to the left of the label (instead of right), there will be a line or box, and the values and labels will be aligned horizontally. See Fig.~\ref{fig:rel_examples} (f)
    \item Below: The label will be below the value (instead of above), with an single line separating the value and label. See Fig.~\ref{fig:rel_examples} (g)

\end{itemize}

The value will be randomly to the right of or below the label for a set (except for To Right, To Left, and Below). If the values are to the right, it is randomly choosen whether they will align horizontally (they start at the same x-position), or not (except in To Right and To Left when it is always aligned).

When placing label-value pairs, there is a probability (which increases with the number of pairs in the column) to start a new column (if there is room horizontally to due so). If the column reaches the bottom of the image a new column is started, unless there isn't horizontal room, in which case the generation of the label-value set ends.

If the label-value set has a header is is either placed at the top-left corner or the top-middle of the label-value set, having a 50\%/50\% chance. If the header is going to be placed at the top-left corner, it has a 50\% of having the label-value pairs begin after it's horizontal position (instead of it being above them).

The placing of the text is done in largely the same manner as the synthetic Wikipedia text.


\emph{Table: \ } 
There is a 33\% chance a header is added for the table. This is 1 to 6 random, non-stop words\footnote{We use the stop words listed at \url{https://www.ranks.nl/stopwords}} from Wikipedia.
A random font and text height are selected for the header, the row and column headers, and the cell text.
A random number of rows (range $[2,15]$) and columns (range $[2,10]$) are selected.
For each header, a length is selected: 81.4\% one word, 18.6\% two words, 6.9\% three words, 2\% four words. That number of non-stop words are randomly sampled from Wikipedia and appended together to form the header.
For each cell, 50\% of the time it will be a single non-stop word sampled from Wikipedia, the other 50\% will be a number with one of the following ten formats (uniformly sampled) displayed in Table~\ref{tab:numbers}.

\begin{table}[h]
\centering
\caption{Number formats for table cells}\label{tab:numbers}
\begin{tabular}{lr}

Description                      & Example \\ \hline
Integer in range [0,100]         & 16     \\
Integer in range [0,9999]        & 4567    \\
Integer in range [-999,999]      & -453    \\
Percent                          & 45\%    \\
Percent with decimal             & 23.45\% \\

Decimal in range [0,100)    & 15.87\% \\

Decimal in range [0,1)           & 0.834   \\
Negative decimal in range (-1,0] & -0.452  \\
Dollar amount in range [0,9999]  & \$2567   \\
Dollar amount in range [0,999]   & \$754    \\

\end{tabular}
\end{table}

The headers and cells are then arranged in a table with some random spacing. The row headers are always on the right of the cells and the column headers are always above the cells.
We leave 15\% of cells blank.
If the table exceeds the space available, the table generation fails (this protects against a bias towards having tables with few rows/columns).

We then draw the lines of the table. All lines have random thickness.
There is always a line separating the headers from the cells. It is randomly deterimined to draw lines between cells and on the outside of the table. Each line is randomly placed (not parallel) in the space availble between the table elements.

\subsubsection{Training tasks}

We define several tasks for these forms, however the Parse to JSON tasks is the most important, as this is also an end task we evaluate on the FUNSD~\cite{funsd} and NAF~\cite{davis} datasets.
We will first detail our JSON format and then list all the tasks.

The JSON format was specifically designed to be easy for an autogressive model to predict. The format must capture the FUNSD dataset labeling, including classes and relationships, in addition to tables which we predict differently.

In general, an instance is represented as a single JSON object: 
\begin{quote}
\texttt{\{"entity text": "class"\}}
\end{quote}
This allows the model to read the text before deciding the class, and during training ensures the model is predicting the class for the right entity. 
If a header has links to other entities, they are listed as \texttt{contents}, e.g.:
\begin{quote}
\texttt{\{"Title Text": "header", "contents":[\{"Q1": "question"\}, \{"Q2": "question"\}]\}}
\end{quote}
If an answer has links, these are handled as \texttt{answers}, e.g.: 
\begin{quote}
    \texttt{\{"Question text": "question", "answers":["A1", "A2"]\}}
\end{quote} We list the answers as strings instead of objects as they should have nothing linked below them in the hierarchy and this is a more compact representation.
Tables are an object with \texttt{row headers}, \texttt{column headers}, and \texttt{cells}, where the cells are a nested list in row major order, e.g.:
\begin{quote}
\texttt{\{"row headers":["R1", "R2"], "column headers":["C1", "C2"], \\"cells":[["r1 c1", "r1 c2"], ["r2 c1", "r2 c2"]]\}}
\end{quote}

We write out the elements in read order, treating a table, or a header with all its sub-elements as a single element.
The read order is determined by first ordering the elements by verticle position. We then take the top element and find all other elements which fall inside a horizontal range slightly above and below it. This is intended to be elements on roughly the same horizontal line, taking large elements (like tables) into account (lots of things can be parallel to them). If the current element is the left most, it is the in order, otherwise, the elements to its left as place before it and they are evaluated with their own horizontally parallel elements. This process makes the read out be roughly natural for how a human might read around blocks like tables.

We note, it would be more efficient and probably more accurate to have defined special tokens for the control characters of the JSON, but we did not do this.

Here is the list of all tasks used in training on the form images:
\begin{itemize}
\item (48.2\%) Parse to JSON: The document is reproduced in a special JSON format which captures structure as well as the class of thee entities. Examples of the JSON can be seen in Fig.~\ref{fig:synthform1} and ~\ref{fig:synthform2}. There are two possible queries, one to parse the document from the beginning, the other includes some portion of the JSON in the query and the model must parse starting from that point of the JSON (similar to the Read On task). This is neccesary as many forms have a JSON longer than the model's longest output (800 tokens)



\item (4.02\%) Link All: The query contains a form entity either by text, highlight, or both, and the model is to predict the class of the entity and read the text of all entities it is linked to.
\item (4.02\%) Link Down: Same as the above task, but only read text of linked entities down the hierarchy
\item (4.02\%) Link Above: Same as the above tasks, but only read text of linked entities up the hierarchy
\item (4.42\%) Cell: The query contains the texts of a row and a column header and the model must read the corresponding cell
\item (4.42\%) Row Header: The query contains a cell's text and the model must read the row header
\item (4.42\%) Column Header: Same as the above task by reading the column header
\item (4.42\%) All Row Cells: The query contains text for a row header and the model must read all the cells in the row.
\item (4.42\%) All Column Cells: Same as above for column
\item (4.42\%) All Row Headers: The query contains a number $i$ and the model must read the row headers for the $i^\textit{th}$ table in the document 
\item (4.42\%) All Column Headers: Same as above for columns
\item (3.61\%) Count Tables: The model must return the number of tables and predict a mask covering them.
\item (4.42\%) Highlight Table: The query contains a number $i$ and the model must predict a mask for the $i^\textit{th}$ table
\item (0.402\%) Not Present: One of the above tasks with a specific query is given, but the entity in the query isn't on the document. The model must respond with a not-present token
\item (0.402\%) Read On: The query as some text and the model must read on from that text to the end of the entity it belongs to
\end{itemize}


\subsection{Selection of ``Easy'' Fonts for Distillation}

We score each font by rendering the following strings in the font: ``abcdefg'', ``hijklmn'', ``opqrst'', ``uvwxyz'', ``12345'', ``67890'', ``ABCDEFG'', ``HIJKLMN'', ``OPQRST'', ``UVWXYZ'' 
We then run Tesseract over on these images and compute the edit-distance between the Tesseract output and the image's source string. The sum of these edit-distances become the score for that font. All fonts with a score less than 21 are used as our ``easy'' fonts. This may seem like a high threshold, but the word images passed to Tesseract are not padded (text generally extends to the end of the image) which is a domain that Tesseract struggles with.

\begin{figure}[pt]
\centering
\includegraphics[width=0.99\textwidth]{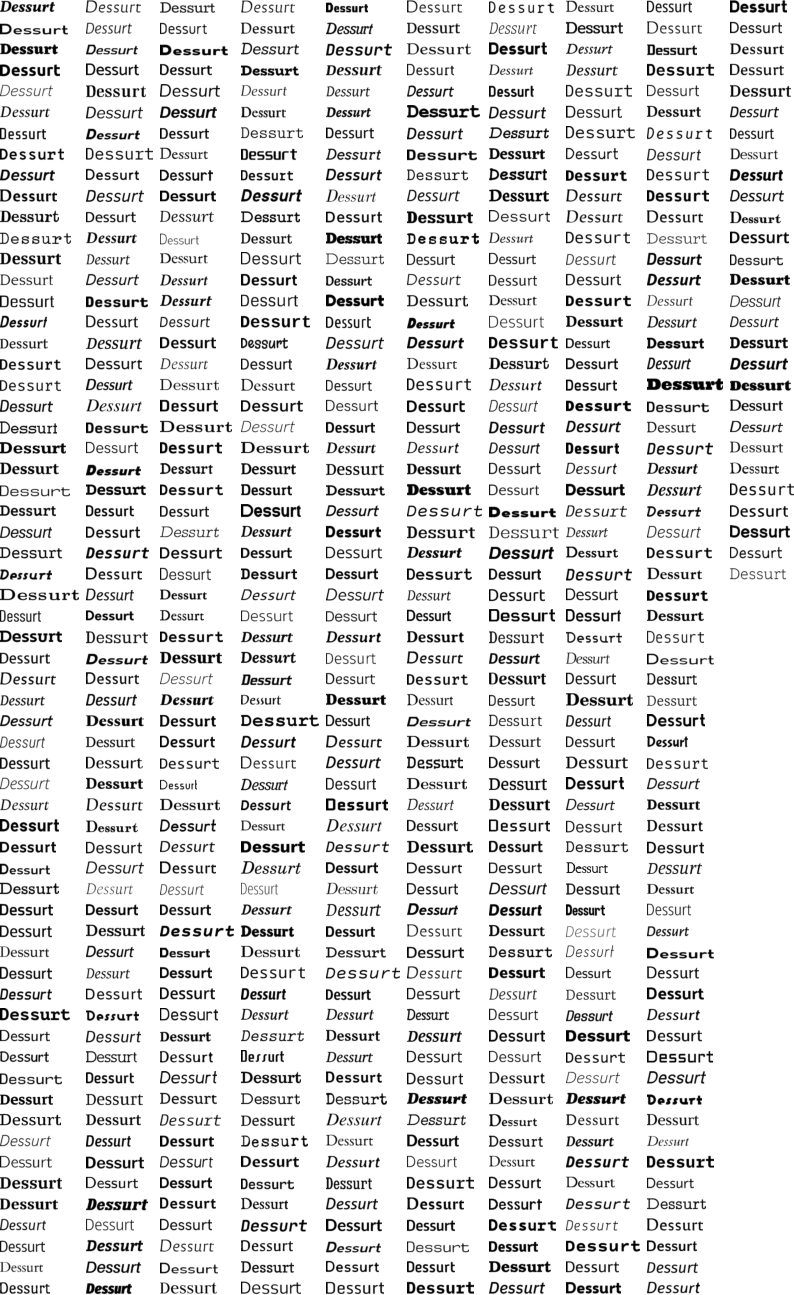}
\caption{All ``easy'' fonts rendering the word ``Dessurt'' at a text height of 16}
\label{fig:easy}
\end{figure}

There are 586 fonts in our ``easy'' set, and they can be seen in Fig.~\ref{fig:easy}.

\subsection{Pre-training Curriculum Details}

It has been noted by \cite{li2021curriculum} that (billion parameter) autoregressive models have training stabilized by a sequence length based curriculum.
This may be related to the success of our curriculum.

The small image pre-training uses the following tasks with uniform probability:
\begin{itemize}
    \item Get Blanked
    \item Re-read Replaced
    \item Highlight Text
    \item Read Highlight
    \item Read On
\end{itemize}

The reading pre-training on full sized images uses the same tasks as normal training, but with uniform probability. As there are more reading focused tasks, this step of the pre-training is focused on teaching reading.

During the main pre-training, the datasets are not sampled uniformly. We assume that some are more important than others. For the final model they are sampled with the following frequency:
\begin{itemize}
    \item IIT-CDIP: 45\%
    \item Synthetic Wikipedia: 29\%
    \item Synthetic Handwriting: 1\%
    \item Synthetic Forms: 5\%
    \item Distillation: 20\%
\end{itemize}

For the ablation experiments, the models using all datasets has the given frequencies, with all others having the same ratio between the datasets they do have:
\begin{itemize}
    \item IIT-CDIP: 53\%
    \item Synthetic Wikipedia: 35\%
    \item Synthetic Handwriting: 2\%
    \item Synthetic Forms: 7\%
    \item Distillation: 2\%
\end{itemize}

The changed frequencies used for the final model reflect the uncertainty the ablation showed regarding the importance of the distillation.




\section{Data Augmentation}

Images of all datasets are scaled to match the size of Dessurt's input.
In all our pre-training and fine-tuning, we apply some basic image augmentations. For most datasets, we randomly re-scale the image to 0.9-1.1 its original size, sampled uniformly. The exception is the census data which is scaled in the range [1,1.15], and the FUNSD dataset, which has the range [0.8,1.2]. If the image is larger than Dessurt's input size (due to a re-scale), we randomly crop a region form the image of Dessurt's input size.
We apply a random rotation from the normal distribution with a standard deviation of $1^{\circ}$.

We apply brightness augmentation which adjusts the brightness and contrast between background and foreground for the synthetic handwriting, synthetic forms, and IAM full-page recognition. The method is the same as used by Tensmeyer et al.~\cite{tensmeyer2017}, but we use $\sigma = 20$.

\section{GAnTED}

Here we describe the greedy-aligned normalized tree edit-distance (GAnTED), the metric we use to evaluate form parsing.

This is simply a greedy optimization of the nTED~\cite{nted} metric done by permuting child lists of the predicted tree. This is neccsary as there is not a canonical ordering for forms. While we create the parse JSONs in a read order, it can often appear ambiguous which elements should be read first. Additionally, the order of the elements should be irrelevant to the information extracted.

The process we use is quite simple, if somewhat slow. We first convert the JSON into a tree. We discard class information in this process. A header will have it's content as children, and a question will have it's answer(s) as a child/children. For tables, things are handed a bit differently. We could have the list of cells in each row be children of its respective row header (the column headers have no children), or have the columns of cells be children of the column headers (the row headers having no children). While the model is trained to predict row-major tables, we note that often errors are made where a table is not recognized as such, and thus the header-cell relationships are predicted instead. Our table annotation of the FUNSD dataset is heuristic (see Section~\ref{sec:tables}) and sometimes erroneous leading to such label-value relationships in the GT. Thus we compute the GAnTED for all combinations of table-to-tree conversions and take the minimum score. 

We use the variant of TED where the relabel cost for the nodes is the normalized Levenshtein distance between the predicted string and the GT string. This means the recognition errors should be balanced in relation to structure errors.

The alignment is done in a breadth first traversal of the predicted tree. At each node, we compute the nTED for the entire tree when the node is moved up to 10 positions forward or backward in its list of children. We then place it in the position that gave the minimum score. Each node gets re-positioned once in this process.
After each node is re-positioned, the final nTED score is the GAnTED score.

This is clearly not optimal, but given that the model attempts to predicted in read order, it is quite stable, only changing the GAnTED slightly if the alignment is done again.

In Table~\ref{tab:ganted} we show the the nTED score, GAnTED score, and the GAnTED score when the alignment is done twice. As can be seen, the greedy alignment dramatically improves the nTED score, likely giving much accurate measures of a model's performance at form parsing, not just how well it matches the order of the GT. We also evaluate computing GAnTED on \modelName{}'s results when each set of children in it's tree are randomly permuted. This leads to decreased performance and less stability, indicating that an approximate read order should be established before computing GAnTED. We feel this should be reasonably easy to do under most situations.

\begin{table}[t]
\caption{nTED, GAnTED, GAnTED with two aligment passes on the FUNSD and NAF datasets}\label{tab:ganted}
\begin{tabular}{l|rrr|rrr}
                                & \multicolumn{3}{c|}{FUNSD}                                   & \multicolumn{3}{c}{NAF}                                      \\
 &
  \multicolumn{1}{l|}{nTED} &
  \multicolumn{1}{l|}{GAnTED} &
  \multicolumn{1}{l|}{2-GAnTED} &
  \multicolumn{1}{l|}{nTED} &
  \multicolumn{1}{l|}{GAnTED} &
  \multicolumn{1}{l}{2-GAnTED} \\ \hline
FUDGE~\cite{fudge} w/ Tesseract            & \multicolumn{1}{r|}{59.1} & \multicolumn{1}{r|}{34.8} & 34.5 & \multicolumn{1}{r|}{-}    & \multicolumn{1}{r|}{-}    & -    \\
  Dessurt (scrambled) &
  \multicolumn{1}{r|}{81.4} &
  \multicolumn{1}{r|}{35.8} &
  \multicolumn{1}{r|}{32.0} &
  \multicolumn{1}{r|}{-} &
  \multicolumn{1}{r|}{-} &
  \multicolumn{1}{r}{-} \\
Dessurt                         & \multicolumn{1}{r|}{44.1} & \multicolumn{1}{r|}{23.4} & 23.2 & \multicolumn{1}{r|}{80.4} & \multicolumn{1}{r|}{42.5} & 42.1 \\
Dessurt w/ census train & \multicolumn{1}{r|}{-}    & \multicolumn{1}{r|}{-}    & -    & \multicolumn{1}{r|}{73.0} & \multicolumn{1}{r|}{38.8} & 38.3
\end{tabular}
\end{table}

\section{Experiment Details}

For each dataset we fine-tune the long-pre-trained model with a learning rate drop and early stopping based the validation set. We took the parameters with the best validation set performance as the final model.

\subsection{RVL-CDIP}

This dataset has significantly more data than the others we evaluate on. We drop the learning rate at 175K iterations, but are able to continue training to a total of 1.5 million iterations with continuous improvement on the validation set.

\subsection{DocVQA}

For DocVQA, we drop the learning rate at 200K iterations and evaluate the model at 380K iterations.

\subsection{HW-SQuAD}

For HW-SQuAD, we drop the learning rage at 200K iterations and evaluate the model at 970K iterations.

\subsection{FUNSD and NAF}

The model frequently falls into the common autoregressive degeneration of repeating the same output (generally a JSON object). We counter this by post-processing the output and removing any sequence of at least 8 characters that is repeated consecutively at least 5 times.
If the model fails to produce the end token, we use the last predicted tokens to form a new query for the model to parse from. We note that this can allow the model to recover from a repeat degeneration, as often it will continue repeating till the maximum token length, these are removed, and then a new query is made from the end of the non-degenerate prediction. We re-query a maximum of 5 times.

Despite the highly regular structure of our JSON output, the model often fails to produce valid JSON output, especially on more difficult forms.
We craft a series of rules to transform various JSON syntax errors into valid JSON, generally favoring a simple, less structured, output. We assume our correction rules don't effect performance significantly as these are generally occurring where the model is making other prediction errors.

During training on the FUNSD~\cite{funsd} and NAF~\cite{davis} datasets, we use the same task distribution as the pre-training on synthetic forms.
While it may not seem intuitive to training on tasks that are not part of the evaluation, the non-JSON tasks do improve performance, possibly providing a regularizing effect.
For the FUNSD dataset, we drop the learning rate at 10K iterations and evaluate the model at 51K iterations.
For the NAF dataset, we drop the learning rate at 65K iterations and evaluate the model at 320K and 400K iterations for the normal and census pre-trained model respectively.

\subsubsection{Table annotations for FUNSD}\label{sec:tables}
The FUNSD dataset doesn't contain annotations for tables. However, tables generally show up distinctively in the annotation with values having two labels linked to them. We use this along with various spatial heuristics to determine if a set of links actually comprise a table. It is generally successful, failing on tables where the label-value linking was left incomplete in the FUNSD annotations.

\subsubsection{Table annotations for NAF}\label{sec:naftables}
The NAF dataset does contain table annotations, however, the transcriptions for the cells is not present in the dataset. We simply omit the \texttt{cells} of the JSON so only row and column headers are predicted. This follows in line with \cite{fudge}, which omits tables from it's predictions.

\subsubsection{U.S.A. 1940 Census pre-training}

\begin{figure}[pt]
\centering
\includegraphics[width=0.99\textwidth]{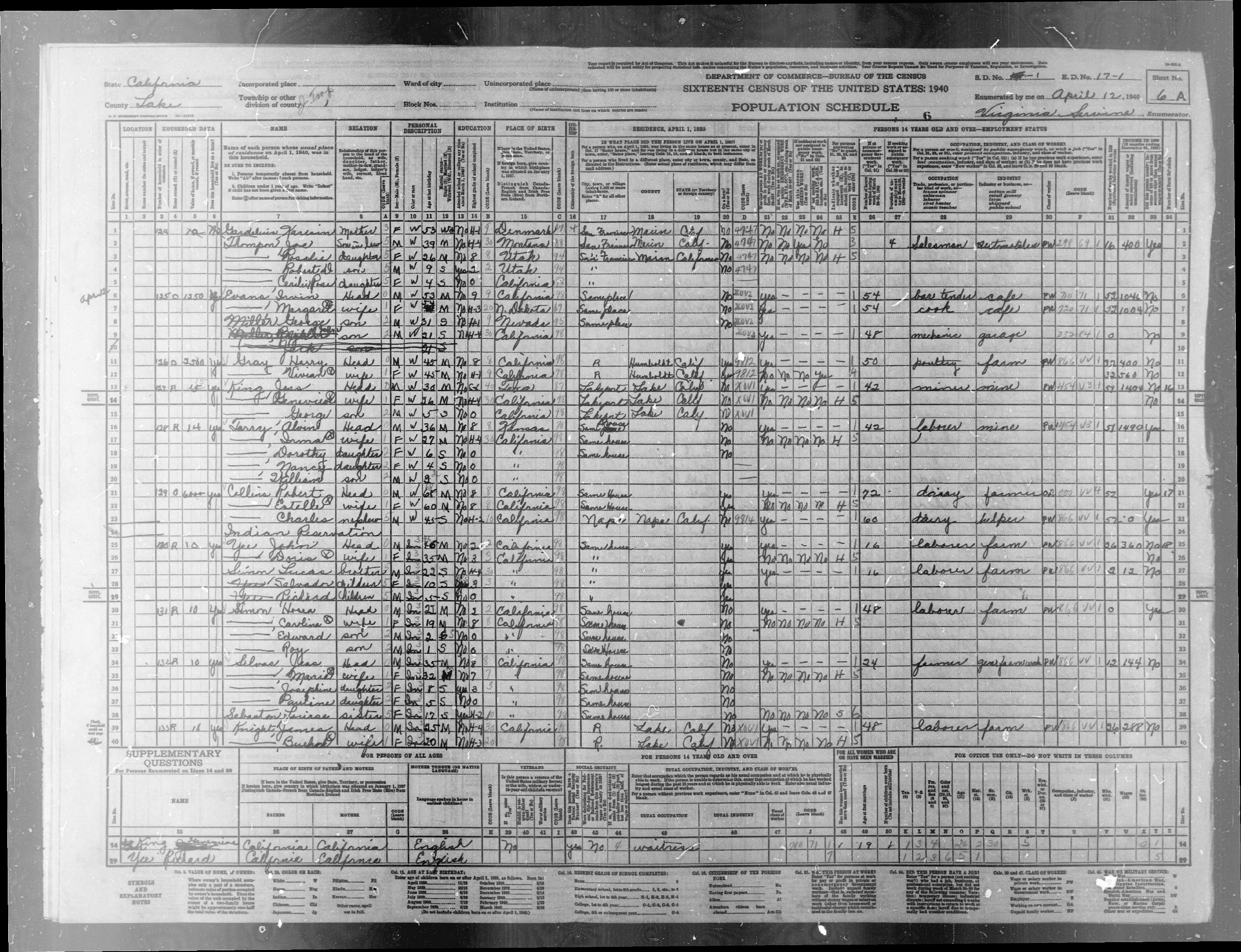}
\caption{Example image from the U.S.A. 1940 census}
\label{fig:census}
\end{figure}

The census images we pre-train on are publicly available on the U.S.A. National Archive at \url{https://www.archives.gov}. The training set we use is 10,000 images. And example image is found in Fig.~\ref{fig:census}. 
The NAF dataset was also derived from the U.S.A. National Archive and thus the census images represents a very similar domain, although they lack any variation in layout. We ensure no overlap between these datasets.

The proprietary annotations we use contain human transcriptions of select columns of the main table in the document: line number, household ID, full name, sex, age, relationship to head of household, race, and birthplace. In our annotations, ditto marks have been filled in with the respective value, and the model is trained to do this as well.
The there are three tasks we use in the pre-training: 
\begin{itemize}
    \item List the full contents of the table, being the above mentioned fields for each row
    \item List all names: List the names on the document. This is the column with the most variation and we assume most handwriting recognition is going to be learned from this column
    \item List all ages: List the ages on the document. Similar to the above task, but ensuring the model can read numbers
\end{itemize}

We crop the images to be only the left-side of the image as the only columns we use are on the left side. This allows the document to fit the aspect ratio of our model better and have higher resolution, which is needed given how dense the handwriting is.

\subsection{IAM Database}

When scoring our IAM NER~\cite{tuselmann} predictions, we do an alignment between the predicted word transcriptions and the GT words, minimizing the total edit-distance. This allows us to match the class prediction even on words the model did not transcribe correctly.

For the experiment pre-training Dessurt on the IAM dataset for IAM NER, for the last 200k iterations of the pre-training, 47\% of the training instances are synthetic documents, each containing two columns of words sampled randomly from three IAM pages (both pages' words are jumbled together). The model must predict the contents of the two columns (full page recognition). By having Dessurt read the words in random order we hope to prevent overfitting on the dataset.
Each word is randomly rescaled to a height in the range of 18 to 48

We note that the IAM splits used for IAM NER are not the same splits used to train the handwriting generation method we used in our data creation~\cite{line_gen}. This means there is a potential information leak of test set data via what the generation model has learned and is using to generate our synthetic pre-trianing data. We feel this would be making a very minor impact on performance especially given the \modelName{}'s performance on IAM recognition~\cite{iam}, which does not have information leakage, is roughly the same as the recognition on the IAM NER splits.

\clearpage
%
%
\bibliographystyle{splncs04}
\bibliography{egbib}